\documentclass[letterpaper]{article} % DO NOT CHANGE THIS
\usepackage{aaai25}  % DO NOT CHANGE THIS
\usepackage{times}  % DO NOT CHANGE THIS
\usepackage{helvet}  % DO NOT CHANGE THIS
\usepackage{courier}  % DO NOT CHANGE THIS
\usepackage[hyphens]{url}  % DO NOT CHANGE THIS
\usepackage{graphicx} % DO NOT CHANGE THIS
\usepackage{natbib}  % DO NOT CHANGE THIS AND DO NOT ADD ANY OPTIONS TO IT
\usepackage{caption} % DO NOT CHANGE THIS AND DO NOT ADD ANY OPTIONS TO IT
\usepackage{algorithm}
\usepackage{algorithmic}

\newcommand{\ziyue}[1]{\textcolor{purple}{\textit{\textbf{Ziyue}: #1}}}
\newcommand{\re}[1]{{\color{black}#1}}

\usepackage{amssymb}
\usepackage[utf8]{inputenc} % allow utf-8 input
\usepackage[T1]{fontenc}    % use 8-bit T1 fonts
\usepackage{url}            % simple URL typesetting
\usepackage{booktabs}       % professional-quality tables
\usepackage{amsfonts}       % blackboard math symbols
\usepackage{nicefrac}       % compact symbols for 1/2, etc.
\usepackage{microtype}      % microtypography
\usepackage{xcolor}         % colors
\usepackage{graphicx}
\usepackage{multirow}
\usepackage{array}
\usepackage{tabularx}
\usepackage{soul}
\usepackage{amsmath}
\usepackage{caption}
\usepackage{subfigure}
\usepackage{booktabs}
\usepackage{threeparttable}
\usepackage{xcolor}


\usepackage{newfloat}
\usepackage{listings}
\DeclareCaptionStyle{ruled}{labelfont=normalfont,labelsep=colon,strut=off} % DO NOT CHANGE THIS
\floatstyle{ruled}
\newfloat{listing}{tb}{lst}{}
\floatname{listing}{Listing}
\title{TimeCMA: Towards LLM-Empowered Multivariate Time Series Forecasting via Cross-Modality Alignment}

\author {
    Chenxi Liu\textsuperscript{\rm 1},
    Qianxiong Xu\textsuperscript{\rm 1}\thanks{Corresponding author},
    Hao Miao\textsuperscript{\rm 2},
    Sun Yang\textsuperscript{\rm 3},
    Lingzheng Zhang\textsuperscript{\rm 4},
    Cheng Long\textsuperscript{\rm 1},\\
    Ziyue Li\textsuperscript{\rm 5}$^*$,
    Rui Zhao \textsuperscript{\rm 6}
}
\affiliations {
    \textsuperscript{\rm 1}S-Lab, Nanyang Technological University, Singapore\\
    \textsuperscript{\rm 2}Aalborg University, Denmark\\
    \textsuperscript{\rm 3}Peking University, China\\
    \textsuperscript{\rm 4}HKUST (Guangzhou), China\\
    \textsuperscript{\rm 5}University of Cologne, Germany\\
    \textsuperscript{\rm 6}SenseTime Research, China\\
    \{chenxi.liu, qianxiong.xu, c.long\}@ntu.edu.sg, haom@cs.aau.dk,
      2201210484@stu.pku.edu.cn, \\ lingzhengzhang01@gmail.com, zlibn@wiso.uni-koeln.de, zhaorui@sensetime.com
}

\usepackage{bibentry}
\begin{document}
\maketitle
\begin{abstract}
    Multivariate time series forecasting (MTSF) aims to learn temporal dynamics among variables to forecast future time series. Existing statistical and deep learning-based methods suffer from limited learnable parameters and small-scale training data. Recently, large language models (LLMs) combining time series with textual prompts have achieved promising performance in MTSF. However, we discovered that current LLM-based solutions fall short in learning \textit{disentangled} embeddings. We introduce TimeCMA, an intuitive yet effective framework for MTSF via cross-modality alignment. Specifically, we present a dual-modality encoding with two branches: the time series encoding branch extracts \textit{disentangled yet weak} time series embeddings, and the LLM-empowered encoding branch wraps the same time series with text as prompts to obtain \textit{entangled yet robust} prompt embeddings. As a result, such a cross-modality alignment retrieves \textit{both disentangled and robust} time series embeddings, ``the best of two worlds'', from the prompt embeddings based on time series and prompt modality similarities. As another key design, to reduce the computational costs from time series with their length textual prompts, we design an effective prompt to encourage the most essential temporal information to be encapsulated in the last token: only the last token is passed to downstream prediction. We further store the last token embeddings to accelerate inference speed. Extensive experiments on eight real datasets demonstrate that TimeCMA outperforms state-of-the-arts.
    %Finally, we develop a \textbf{time series forecasting} module to decode the aligned embeddings while capturing dependencies among multiple variables for MTSF. 
\end{abstract}

\begin{links}
    \link{Code}{https://github.com/ChenxiLiu-HNU/TimeCMA}
\end{links}

\section{Introduction}
\label{introduction}

% ========================================
% Paragraph 1
% Background
% ========================================

\if 0
\ziyue{thoughts from ziyue:\\
- existing text-time dual modality LLM for prediction: directly concat or [blabla]\\
- they have the limitations of [blabla]: (1) ignore multivariate relations --> (2) overlength prompt and thus sacrificing the scalability, (3) redundant info in the LLM's embeddings\\
- our solution: (1) inverse self-attention for multi-variate relation, (2) last token, (3) cross attention for retrieval and purifying the LLM's embedding.\\
\textbf{Conclusion:}\\
- Title: Last Pure Token Is Enough for Scalable Text-Time Series LLM-based Multivariate Prediction.\\ 
- Claim: pure token (entanglement), last token (easing computation burden), multi-variate relation}
\fi

With the proliferation of scalable mobile sensing, large amounts of time series data, collected across domains such as traffic~\cite{DBLP:journals/tits/XiaoX0HLZ022,miao2024unified} and environment~\cite{liu2024mvcar,DBLP:journals/www/LiuXWCCC22}, have driven applications such as multivariate time series forecasting (MTSF). MTSF aims to mine temporal dynamics among variables from historical data to predict future time series, enabling users to make proactive decisions, e.g., investment choice~\cite{DBLP:journals/eswa/NiuWLYD20} or weather preparation~\cite{DBLP:journals/corr/abs-2310-06625}.
%In this way, users can make decisions in advance, for example, weather forecasting makes it possible to better prepare for severe weather conditions~\cite{DBLP:journals/corr/abs-2402-03784}.

% ========================================
% Paragraph 2
% Existing MTSF models and its limtations.
% ========================================
%\ziyue{unnecesary, too detailed like related-work, we }
% statistical + shallow
% ========================================
% Paragraph 3
% LLM-based models (Time series-based models and prompt-based models) -> the latter is better
% ========================================
% Limitation of existing MTSF:
%   a) small amount of parameters.
%   b) witness only 1 small dataset.
%   => suboptimal embeddings/performance.
% 
% Introduce merits of LLM and briefly introduce some existing LLM for MTSF.
% ========================================
% llm
%\ziyue{the existing LLMs are not for MTSF; we are the first one, so this is also an important contribution, and it refers to our variate attention.} 
MTSF methods can be divided into statistical methods~\cite{smith1997traffic} and deep learning-based methods~\cite{wu2023timesnet,miao2022mba}. However, the limited number of learnable parameters and the small-scale training data prevent these methods from achieving better performance and being more robust~\cite{DBLP:journals/corr/abs-2402-02713,21-TNSE-Liu,chenxi2021study}. 
Recent advances~\cite{DBLP:conf/nips/ZhouNW0023} have incorporated pre-trained LLMs~\cite{radford2019language, radford2018improving} into time series to benefit from the \textit{robust} embeddings learned from abundant language data~\cite{liang2024foundation}.

Existing LLM-based methods for time series forecasting can be categorized by input data types. (1) \textit{Time series-based LLMs}~\cite{DBLP:conf/nips/ZhouNW0023,liu2024spatial} replace the LLM's tokenizer with a randomly initialized embedding layer to process time series data. However, this embedding layer initialized with random weights often results in weak embeddings due to a domain gap between time series and language data.
(2) \textit{Prompt-based LLMs}~\cite{DBLP:journals/corr/abs-2402-02713} introduce prompts with text as additional input to help the LLMs understand the time series forecasting task. The prompts are contextualized within time series with text descriptions to facilitate the data-to-text transformation~\cite{jin2023time,xue2023promptcast} or directly summarize the time series information using pure text~\cite{liu2024unitime,DBLP:conf/aaai/JiaWZCL24,huangleret}. These prompts are then processed by pre-trained LLMs to obtain robust embeddings, allowing the prompt-based LLMs to outperform existing methods, as evidenced in Table~\ref{tab:main}.

\begin{figure}[t]
    \begin{centering}
    % \vspace{-0.5cm}
\includegraphics[width=\columnwidth]{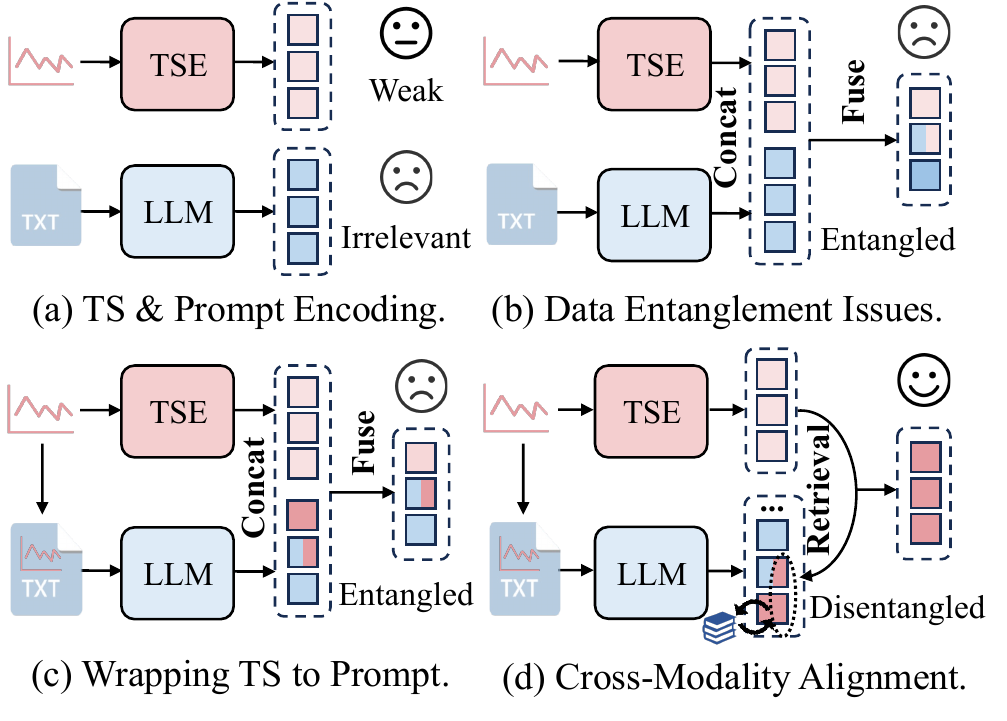}
        % \vspace{-10pt}
        \caption{\small (a) Limits of single-modality models: time series encoder (TSE) offers disentangled yet weak embeddings (in light red); text-only model learns textual embeddings (in blue), irrelevant to time series. (b) Existing models directly fuse two modal embeddings, leading to data-entangled issues. (c) Some tried to wrap time series into prompt, enhancing temporal component in prompt embedding, yet still yielding entanglement. (d) Our method obtains disentangled and robust time series embedding (dark red) via similarity-based retrieval, with last token embeddings stored for efficient forecasting.} 
        \label{fig:intro}
    \end{centering}
    % \vspace{-0.6cm}
\end{figure}

% ========================================
% Paragraph 4
% Limitation of prompt-based LLM for MTSF:
%   a) although high-quality embedding can be obtained, the time series information is mingled with redundant sentence information, which is useless for MTSF.
% ========================================
%\ziyue{only explained one challenge; maybe add one demo figure to show the entanglement issue and last token.} 
 
Although prompt-based LLMs have achieved notable performance, they were challenged by the \textbf{\textit{data entanglement issue}}~\cite{chang2024timedrl}. Specifically, existing methods~\cite{liu2024unitime,DBLP:conf/aaai/JiaWZCL24} concatenate \textit{disentangled yet weak} time series embeddings (Fig. \ref{fig:intro}(a), upper) with text prompt embeddings (Fig. \ref{fig:intro}(a), lower). As one stream of existing methods shown in Fig. \ref{fig:intro}(b), these fused embeddings are then fed into subsequent time series processing stages. However, the output embeddings are entangled, which degrades the forecasting performance because the textual information acts as noise. How to potentially mitigate the noisy textual embedding? As in Fig. \ref{fig:intro}(c), one attempt is to wrap time series values within text prompts, strengthening the time series embeddings while retaining text to enable LLMs to better understand the time series information as natural language~\cite{DBLP:conf/gis/0001VS22}. Nevertheless, due to the nature of the concatenation method and Transformer blocks within the LLMs, the prompt embeddings become \textit{entangled yet robust}, leading to sub-optimal performance, as in Fig. \ref{fig:ab_model}. To address this challenge, we propose that only the \textbf{\textit{disentangled and robust time series embeddings}} from LLMs are optimal for MTSF, and this can be easily achieved by our intuitive cross-modality alignment design via similarity-based retrieval for enhancing forecasting, as in Fig. \ref{fig:intro}(d).

% ========================================
% Paragraph 5
% high-level idea - core of this paper.
% we propose a dual branch model to:
%   a) use a small-scale model to extract relatively low-quality embeddings (pure time series).
%   b) wrap the time series a a prompt, and use prompt-based LLM to obtain high-quality embeddings (time series + words).
%   c) employ a cross attention, using low-quality time series embeddings to retrieve relevant high-quality time series embeddings, to augment the former.
% Exp: Table 2 main results and Figure 3 ablation study
% ========================================

Overall, we present an LLM-empowered framework for multivariate \underline{time} series forecasting via \underline{c}ross-\underline{m}odality \underline{a}lignment, called \textbf{TimeCMA}. It has a dual-modality encoding module with a time series encoding branch and an LLM-empowered encoding branch, a cross-modality alignment module, and a time series forecasting module.
%The time series encoding branch extracts \textbf{pure} embeddings from historical time series data. At the same time, the LLM-empowered prompt encoding branch wraps the same time series as text prompts for obtaining \textbf{high-quality} yet \textit{entangled} embeddings. 
The time series encoding branch extracts variable embeddings from historical time series data. The LLM-empowered prompt encoding branch wraps the same time series as prompts to obtain embeddings with well-trained knowledge.
Then, the cross-modality alignment module is designed to integrate the two groups of embeddings. Intuitively, as in Fig.~\ref{fig:last_token}, the time series embeddings (light red) would naturally have stronger \re{correlations} with the time series \re{component} (dark red) in the prompt embeddings (mixed color). Therefore, the robust time series \re{components} are retrieved from the prompt embeddings based on channel-wise similarity and then aggregated into the original ones (light red) to enhance forecasting. 

% ========================================
% Paragraph 6
%  Further issues for prompt-based LLM:
%   a) Heavy training costs 
%   b) Slow inference speeds
% ========================================
Nonetheless, prompt-based LLMs \re{suffer from} high computational costs and slow inference speeds \re{because:}
\textit{(i) The characteristics of multivariate time series (MTS) data}: unlike 1D prompt data (with $N$ tokens), MTS data has two dimensions: variable and time (with $N$ variables and $T$ timestamps), causing a substantial computational load.
\textit{(ii) High computational burden of LLM outputs.} Despite attempts to reduce computational costs by freezing partial or all of the LLM’s parameters, prompt-based LLMs remain computationally expensive because multi-head attention in LLMs generate high-dimensional outputs and require substantial computational power.
\textit{(iii) Repetitive processings with the frozen LLMs}: during training, existing prompt-based LLM~\cite{jin2023time} performs online processing with the frozen LLMs. Consequently, each training sample is processed repetitively by the LLM in each training epoch, though the obtained embeddings remain unchanged due to the frozen parameters. Therefore, the inference speed is considerably slower.
% \re{is this a general problem of LLM-based MTSF or is it just this paper? if latter, we should delete (iii)}. 

% ========================================
% Paragraph 7
% Further design:
%   a) prompt
%   b) last token embedding tailor
% Exp: Figures 4 and 5 prompt design; and Table 4 efficiency
% ========================================
\re{To ease the computational burden}, we further propose the last token embedding storage. \re{\textbf{(1) The last token is enough}: in the prompt, we independently} wrap time series data of each variable to preserve the characteristics of MTS data; then, \re{we tailor the prompt design so that the LLM is instructed to encapsulate vital temporal essences} into the last token of each prompt.
By only feeding this embedding to align with the time series, we can reduce the computational cost.
\re{\textbf{(2) Offline storage}:} we store the last token embedding to avoid repetitive processing with frozen LLM, thereby accelerating the inference speed. 
%Finally, the time series forecasting module decodes the aligned embeddings while capturing dependencies among variables for MTSF.
% ========================================
% Paragraph 8
% Contribution
% ========================================
%TODO
Our contributions are:
% \vspace{-0.1cm}
\begin{itemize}
    %\item We propose an efficient LLM-empowered multivariate time series forecasting framework named TimeCMA, {\cheng which} integrates a time series branch and a prompt branch in a dual-modality module to capture temporal dependencies among multiple variables and {\cheng to} generate {\cheng the} embedding {\cheng for each modality}.
    % , respectively.
    \item \re{We identify data entanglement issues in the embeddings of dual-modality LLMs for time series forecasting and proposed a TimeCMA framework to learn disentangled embeddings from LLM with text-time series data.}
    %\item We propose an LLM-empowered framework named TimeCMA for multivariate time series forecasting, which integrates a time series branch and a prompt branch in dual modality to capture temporal dependencies among multiple variables.
% and {\cheng to} generate {\cheng the} embedding {\cheng for each modality}.
    \item The cross-modality alignment module retrieves disentangled and robust time series embeddings from the LLM-empowered prompt embeddings via channel-wise similarity to enhance forecasting.
    
    \item We tailor the last token of each prompt to reduce the computational costs. We then store these last token embeddings to avoid repetitive processings with the frozen LLM for efficient forecasting.

    \item Extensive experiments on eight datasets demonstrate that TimeCMA outperforms state-of-the-arts.
    % Our study manifests that LLMs can serve as an efficient enhancer for MTSF.
    %Extensive evaluations demonstrate that TimeCMA is a generalizable and efficient time series learner that outperforms state-of-the-art MTSF models. 

\end{itemize}
% \vspace{-0.2cm}

\section{Related Work}
\label{relatedwork}

Deep learning models have shown 
crucial promise in time series forecasting. Convolutional neural networks (CNNs) simultaneously capture variable and temporal correlations~\cite{wu2023timesnet,JIN2022315}, while Transformers excel due to their powerful learning capabilities. Early Transformer-based methods~\cite{haoyietal-informer-2021,DBLP:conf/icml/ZhouMWW0022} treat multiple variables at the same timestamp as a single temporal token, which often leads to suboptimal performance by conflating unrelated variables. PatchTST~\cite{Yuqietal-2023-PatchTST} mitigates this issue with a channel-independent configuration but overlooks inter-variable dependencies, resulting in longer training times and weaker performance on datasets with many variables. iTransformer~\cite{DBLP:journals/corr/abs-2310-06625} addresses these limitations by treating independent time series as tokens to better capture multivariate correlations. Despite these advances, existing deep learning methods remain constrained by limited parameterization and small-scale training data~\cite{DBLP:journals/www/CaiWCLX24,liu2024icde,liu2021understanding,DBLP:conf/hpcc/ChenWL20}.

Recently, large language models (LLMs) have achieved superior performance in time series analysis, benefiting from extensive parameterization and large-scale training data~\cite{DBLP:conf/nips/GruverFQW23,DBLP:journals/corr/abs-2402-02713,DBLP:conf/dexa/YangSLLMLZ24}. LLM-based forecasting methods can be categorized as time series-based or prompt-based, depending on whether prompts are included in the input. Time series-based LLMs fine-tune models for univariate~\cite{DBLP:conf/nips/ZhouNW0023} or multivariate forecasting~\cite{liu2024spatial} by replacing the tokenizer with a randomly initialized embedding layer. However, embeddings trained on limited data often suffer due to the domain gap between time series and language data. To address this, prompt-based LLMs incorporate prompts as full or partial input. Early works~\cite{DBLP:conf/gis/0001VS22,xue2023promptcast} explored pure prompting techniques for time series forecasting. Subsequent studies demonstrated that combining time series with prompts~\cite{jin2023time,liu2024unitime,cao2024tempo} or leveraging pre-trained LLM knowledge~\cite{pan2024s2,DBLP:journals/corr/abs-2308-08241} can enhance performance. However, these approaches still face challenges such as data entanglement and high computational costs.

\section{Preliminaries}
\label{problem}
%In this section, we present some definitions used throughout this paper and then formulate the problem of multivariate time series forecasting.
%\textit{Defnition~\ref{problem}.1} (\textbf{Time Series}). A time series is represented as $\mathbf{X}_T=\left\{\mathbf{x}_1, \ldots, \mathbf{x}_{T}\right\} \in \mathbb{R}^{T \times N}$, where $T$ is the number of timesteps and $N$ is the number of variables. Each $\mathbf{x}_i$ is a $N$-dimensional vector, which indicates the values of $N$ variables at timesteps $t_i$. The time series value of a variable at $t_i$ can be represented as $v_i$, as depicted in Fig.~\ref{fig:prompt5} (a).

\begin{figure*}[!t]
    \begin{centering}	
    \includegraphics[width=\textwidth]{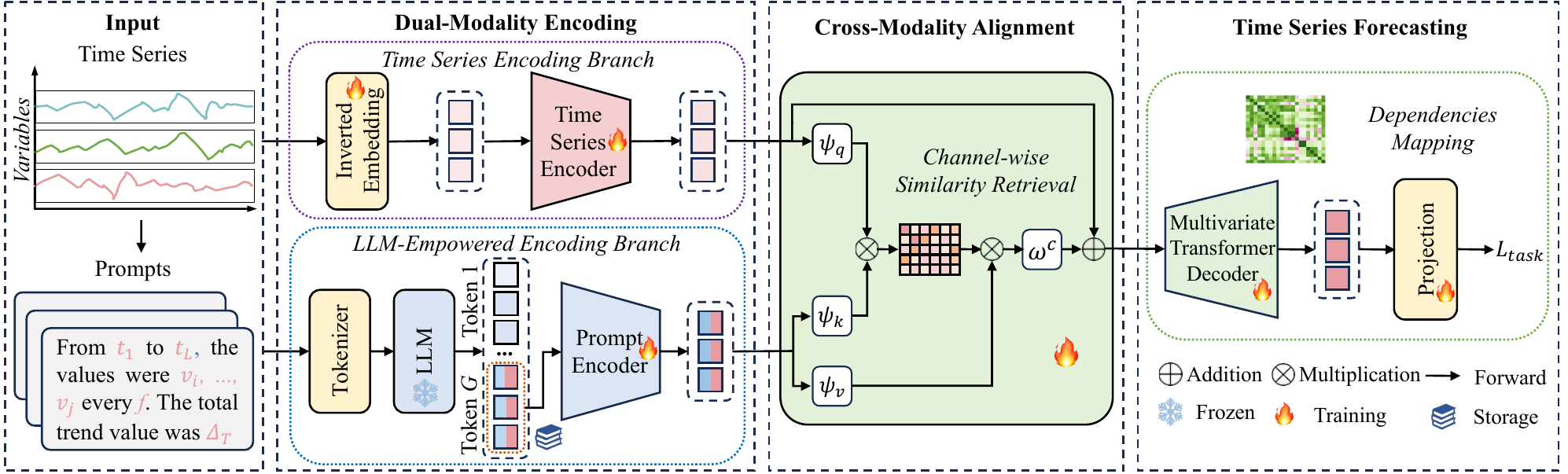}
    \caption{Overall Framework of TimeCMA.}
        % \vspace{-0.5cm}
        % \caption{\small TimeCMA framework. Given time series data and corresponding prompts, we embed them through \textbf{Dual-Modality Encoding}. The last token $G$ of the prompts is tailored, and its embeddings are stored for efficient inference. During \textbf{Cross-Modality Alignment}, LLM-empowered time series embeddings are retrieved from the prompt embeddings based on channel-wise similarity. \textbf{Time Series Forecasting} decodes the aligned embeddings while capturing multivariate dependencies for robust forecasting.}
        %\caption{Overview of TimeCMA. The upper is the \textit{Time Series Modality} to extract embeddings. The lower depicts the \textit{Prompt Modality}, which generates prompt embeddings through frozen LLM. The dual-modality alignment module retrieves time series embeddings from prompt embeddings, and the decoder and projection layer produces the final time series output.
        %\ziyue{the caption should be self-sufficient and self-explained. can be as long as 3-4 lines.}
        %%}
        \label{fig:framework}
    \end{centering}
    % \vspace{-0.5cm}
\end{figure*}

\textbf{Multivariate Time Series}. It is denoted as $\mathbf{X} =\left\{\mathbf{x}_{1}, \ldots, \mathbf{x}_{L}\right\} \in \mathbb{R}^{L \times N}$, where $L$ is the number of time steps and $N$ is the number of variables. 
%Each $\mathbf{x}_l$ is a $N$-dimensional vector, which indicates several $N$ variables (e.g., load or temperature) at time step $l$, where we use $v_i$ to denote the value of $i$-$th$ variable. %We denote the value of $i$-$th$ variable at timestep $t_i$ as $v_i$, as shown in Fig.~\ref{fig:prompt5}(a).

\textbf{Prompt}. We wrap the time series $\mathbf{X} \in \mathbb{R}^{L \times N}$ into prompts $\mathbf{P}_S =\left\{\mathbf{p}_1, \ldots, \mathbf{p}_N\right\} \in \mathbb{R}^{S \times N}$ along with variables, as depicted in Fig.~\ref{fig:framework}. Each prompt $\mathbf{p}_i$ has $S$ elements containing words and time series values.
In the prompt, the $<\textit{italic}>$ elements represent time information, such as timestamps and frequency. The $<\textcolor[HTML]{E69CA1}{color}>$ elements denote time series values of $L$ timesteps. The last value that summarizes temporal information is quantified by the total trend \(\Delta_T\), defined as:
\begin{equation}
    \Delta_T = \sum_{i=1}^{T-1} \delta v_i,
\end{equation}
where $\delta v_i = v_{i+1} - v_i$ represents the incremental change between consecutive timesteps.

%\textit{Defnition~\ref{problem}.2} (\textbf{Prompt}). A prompt is formulated by the time series of a variable with $T$ timesteps, as depicted in Fig.~\ref{fig:prompt5}, denoted as
%$\mathbf{P}_T=\left\{\mathbf{p}_1, \ldots, \mathbf{p}_N\right\} \in \mathbb{R}^{S \times N}$, where each prompt $\mathbf{p}_i$ has $S$ elements containing text and numeric values. Additionally, $T~<~S$. As shown in Fig.~\ref{fig:prompt5} (b), the $<\textbf{bold}>$ elements represent time information, and $<color>$ elements (such as pink) denote time series values. Moreover, each variable with $T$ timesteps corresponds to one prompt. The overall trend within one prompt is quantified by the total trend value \(\Delta_T\), defined as:
%\begin{equation}
%    \Delta_T = \sum_{i=1}^{T-1} \delta v_i,
%\end{equation}
%where \(\delta v_i = v_{i+1} - v_i\) represents the incremental change between consecutive measurements. %This structured prompt formulation encapsulates the contextual and numerical data and precisely delineates the evolving trends over the specified period.

\textbf{Problem Definition.} Given an observation in a multivariate time series $\mathbf{x}_t \in \mathbb{R}^{N}$, where $t$ is a time step.
%and $N$ indicates the number of variables. 
Our goal is to learn a function using historical data $ \mathbf{X}_T =\{\mathbf{x}_{t-T+1:t}\} \in \mathbb{R}^{T \times N}$ with $\mathbf{P}_S$ to forecast future multivariate time series $\widehat{\mathbf{X}}_M = \left\{\widehat{\mathbf{x}}_{t+1:t+M}\right\} \in \mathbb{R}^{M \times N}$ over $M$ timesteps.

%An dual-modality time series forecaster is a function that takes as input as historical time series $\mathbf{X}_T$ and prompts $\mathbf{P}_T$, forecast the future time series $\widehat{\mathbf{X}}_M = \left\{\widehat{\mathbf{x}}_{T+1}, \ldots, \widehat{\mathbf{x}}_{M}\right\} \in \mathbb{R}^{M \times N}$. The forecasting procedure is formulated as follows.
%\begin{equation}
%\mathcal{F}_{\theta}\left(\mathbf{x}_{1}, \ldots, \mathbf{x}_T\right)=\left(\widehat{\mathbf{x}}_{T+1}, \ldots, \widehat{\mathbf{x}}_{M}\right),
%\end{equation}
%\noindent where $\theta$ is the parameters of $\mathcal{F}$.
%\HAO{Given historical multivariate time series $\mathbf{X}_T = \left\{\mathbf{x}_{1:T}\right\}$ over $T$ timesteps and corresponding prompts $\mathbf{P}_S = \left\{\mathbf{p}_{1:N}\right\}$ of $N$ variables, our goal is to learn a function to forecast future time series $\hat{\mathbf{X}}_M = \left\{\hat{\mathbf{x}}_{T+1:M}\right\}$ over $M$ timesteps.}

\section{Methodology}
% We introduce TimeCMA, an LLM-empowered framework for multivariate \underline{time} series forecasting with \underline{c}ross-\underline{m}odality \underline{a}lignment. We first provide an overview of the framework and then introduce specifics on each module.

\subsection{Framework Overview}
TimeCMA contains three key modules: dual-modality encoding, cross-modality alignment, and time series forecasting, as shown in Fig.~\ref{fig:framework}.

\textbf{Dual-Modality Encoding} include a time series encoding branch and an LLM-empowered encoding branch, to effectively learn embeddings for input time series and prompts.
%, respectively. %Through this module, we can separate the time series and text prompt embeddings to enhance more effective feature learning.

\emph{Time Series Encoding Branch} consists of an inverted embedding layer and a time series encoder. The inverted embedding treats an entire variable's time series as a token~\cite{liu2024spatial}, generating token embeddings that are fed into a Pre-LN Transformer encoder~\cite{DBLP:conf/icml/XiongYHZZXZLWL20}.
%the time series encoder is composed of stacked temporal blocks, containing several self-attention layers followed by feed-forward fully-connected layers.}
%Time series data is inputted into the encoding module, consisting of a data embedding layer and a time series encoder. In the data embedding layer, the entire time series of a variable is embedded as a token. The encoder then employs self-attention to capture mutual interactions, followed by individual processing through feed-forward networks to generate the time series embeddings.

\emph{LLM-Empowered Encoding Branch} comprises a frozen LLM, and a prompt encoder with the same architecture as that in the time series encoder. The frozen LLM extracts prompt embeddings with sufficient information extracted from the times series, while the prompt encoder refines these embeddings across multiple variables.
%In existing studies, the multivariate time series with $C$ features is considered as a sequence of multi-channel variables, which often incurs memory overflow, especially in LLM settings~\cite{Yuqietal-2023-PatchTST}. To reduce the computational cost, we propose to convert all variables in a time series into one prompt. We then leverage the pre-trained LLM and a prompt encoder to perform feature extraction for text prompt embeddings, considering the powerful capabilities of LLM for text understanding.
%Time series data encompasses variable and temporal dimensions, leading to heavy computational costs. To address this, we design a prompt for each variable to compress its time series values of timesteps $T$ into prompt tokens. 

\textbf{Cross-Modality Alignment} aggregates the dual modalities. The purpose is to retrieve time series embeddings from the prompt embeddings based on their similarity.

\textbf{Time Series Forecasting} has a multivariate Transformer decoder similar to that in the lightweight Pre-LN Transformer, which decodes the aligned time series embeddings and then inputs them into a projection function for future forecasting.
%The decoder aims to facilitate further long-term temporal dependencies learning between multiple variables. Finally, the decoded embeddings are input into a projection function for future forecasting.
%we use a temporal decoder based on stacked Pre-LN Transformers to decode the embeddings learned from the dual-modality alignment module. The temporal decoder aims to facilitate further long-term temporal dependencies learning. Finally, the decoded embeddings are input into a projection function for prediction.
%The temporal decoder enhances the time series embeddings between variables. These embeddings are delivered to a projection layer to forecast the future time series.

%Next, we will elaborate on the technical details of each module, respectively.

\subsection{Dual-Modality Encoding}
\subsubsection{Time Series Encoding Branch} 
% Previous time series forecasters treat multiple variables of the same timestamp as a temporal token~\cite{DBLP:conf/nips/ZhouNW0023}. These often result in suboptimal performance, as they conflate the representations of multiple unrelated variables. While channel independence addresses this, it leads to longer training times and is unsuitable for data with many variables. 
The time series branch employs an inverted embedding~\cite{liu2024spatial}, which defines the entire time series of a variable as a token, to generate token embeddings. The time series encoder effectively captures complex temporal dependencies between these tokens. 

\textbf{Inverted Embedding.} Given the time series data $\mathbf{X}_T \in \mathbb{R}^{T \times N}$, the inverted embedding aims to convert $\mathbf{X}_T$ into learnable matrices $\mathbf{H}_T \in \mathbb{R}^{C \times N}$ to capture the temporal dependencies of variables~\cite{DBLP:journals/corr/abs-2310-06625}. The $\mathbf{X}_T$ is initially normalized to have zero mean and unit standard deviation via reversible instance normalization to mitigate the time series distribution shift~\cite{DBLP:conf/iclr/KimKTPCC22}. Then, the normalized $\mathbf{X}_T$ is transformed to variable embedding:
\begin{equation}
\mathbf{H}_T = \mathbf{W}_e \mathbf{X}_T + \mathbf{b}_e,
\end{equation}
\noindent where
%$\mathbf{H}_T = \left\{\mathbf{h}_1, \ldots, \mathbf{h}_{T}\right\} \in \mathbb{R}^{C \times N}$ is the output of the embedding layer. 
$C$ indicates the hidden dimension of the embedded time series. $\mathbf{W}_e$ and $\mathbf{b}_e$ are the learnable parameters.

\textbf{Time Series Encoder.} The variable embeddings $\mathbf{H}_T$ are fed into a lightweight encoder $\mathit{TSEncoder(\cdot)}$. Inspired by the Transformer structure in existing LLMs~\cite{xu2024pefad}, we apply layer normalization first in the encoder, meaning it occurs before both the multi-head attention and feed-forward layers. Compared with the original Transformer, this Pre-LN Transformer has the advantages of being more stable and converging faster~\cite{DBLP:journals/pami/HuangQZZLS23}. 
In $\mathit{TSEncoder(\cdot)}$, the embeddings $\mathbf{H}_T^i$ undergo $i_{\text{th}}$ layer normalization $\mathit{LN}(\cdot)$:
\begin{align}
\widetilde{\mathbf{H}}_T^{i} &= \mathit{LN}(\mathbf{H}^{i}_T), \\
% \overline{\mathbf{H}}_T^{i} &= \mathit{MHSA}(\widetilde{\mathbf{H}}_T^{i}) + \mathbf{H}^{i}_T, \\
% \grave{\mathbf{H}}_{T}^{i+1} &= \mathit{LN}(\overline{\mathbf{H}}^{i}_T), \\
% \mathbf{H}_{T}^{i+1} &= \mathit{FFN}(\mathit{LN}(\grave{\mathbf{H}}^{i}_T)) + \overline{\mathbf{H}}^{i}_T, \\
\mathit{LN}(\mathbf{H}^{i}_T) &= \gamma \odot \frac{\mathbf{H}^{i}_T - \mu}{\sigma} + \beta, \label{eq:ln}
% \mathit{MHSA}(\mathbf{H}^{i}_T) &= \rho_o (\mathit{Attention}(\rho_q \mathbf{H}^{i}_T, \rho_k \mathbf{H}^{i}_T, \rho_v \mathbf{H}^{i}_T)), \\
% \mathit{Attention}(\mathbf{H}^{i}_T) &= \mathit{softmax}\left(\frac{\mathbf{H}^{i}_T \mathbf{H}^{iT}_T}{\sqrt{d_k}}\right) \mathbf{H}^i_T,
% \mathit{FFN}(\overline{\mathbf{H}}^i_T) &= \max(0, \mathbf{W}_1\mathbf{H}^i_T + \mathbf{b}_1) \mathbf{W}_2 + \mathbf{b}_2,
\end{align}

\noindent where $\widetilde{\mathbf{H}}^i_T$ represents the intermediate embedding after the $i_{\text{th}}$ layer normalization. $\gamma$ and $\beta$ are learnable scaling and translation parameters. $\mu$ and $\sigma$ represent the mean and standard deviation. $\odot$ denotes element-wise multiplication.

Then, they are processed by the multi-head self-attention mechanism, denoted as $\mathit{MHSA}(\cdot)$. The output, $\overline{\mathbf{H}}_T^{i}$, is combined with $\mathbf{H}_T^i$ through a residual connection: 
\begin{align}
% \widetilde{\mathbf{H}}_T^{i} &= \mathit{LN}(\mathbf{H}^{i}_T), \\
\overline{\mathbf{H}}_T^{i} &= \mathit{MHSA}(\widetilde{\mathbf{H}}_T^{i}) + \mathbf{H}^{i}_T, \\
\mathit{MHSA}(\mathbf{H}^{i}_T) &= \rho_o (\mathit{Attention}(\rho_q \mathbf{H}^{i}_T, \rho_k \mathbf{H}^{i}_T, \rho_v \mathbf{H}^{i}_T)), 
% \\
% \mathit{Attention}(\mathbf{H}^{i}_T) &= \mathit{softmax}\left(\frac{\mathbf{H}^{i}_T \mathbf{H}^{iT}_T}{\sqrt{d_k}}\right) \mathbf{H}^i_T,
% \mathit{FFN}(\overline{\mathbf{H}}^i_T) &= \max(0, \mathbf{W}_1\mathbf{H}^i_T + \mathbf{b}_1) \mathbf{W}_2 + \mathbf{b}_2,
\end{align}

\noindent where $\mathbf{\overline{H}}^i_T$ is output of the $i_{\text{th}}$ layer after the $\mathit{MHSA}(\cdot)$. $\rho_o$, $\rho_q$, $\rho_k$, and $\rho_v$ are the linear projections.

Followed by another $\mathit{LN}(\cdot)$. The normalized $\grave{\mathbf{H}}_{T}^{i+1}$ are then passed through a feed-forward network $\mathit{FFN}(\cdot)$ of fully connected layers that further process the embeddings, then combined with the $\overline{\mathbf{H}}_T^{i}$ through another residual connection:
\begin{align}
\grave{\mathbf{H}}_{T}^{i+1} &= \mathit{LN}(\overline{\mathbf{H}}^{i}_T), \\
\overline{\mathbf{H}}_{T}^{i+1} &= \mathit{FFN}(\grave{\mathbf{H}}^{i+1}_T) + \overline{\mathbf{H}}^{i}_T, 
% \\
% \mathit{FFN}(\overline{\mathbf{H}}^i_T) &= \max(0, \mathbf{W}_1\mathbf{H}^i_T + \mathbf{b}_1) \mathbf{W}_2 + \mathbf{b}_2,
\end{align}

\noindent where $\mathbf{\grave{H}}^{i+1}_T$ represents the intermediate embedding of the $i_{\text{th}}$ layer after the second $\mathit{LN}(\cdot)$. To simplify, $\overline{\mathbf{H}}_T \in \mathbb{R}^{C \times N}$ symbolizes the output of $\mathit{TSEncoder(\cdot)}$.

\subsubsection{LLM-Empowered Encoding Branch}
Pre-trained LLMs learn from input tokens, making them more sample-efficient than encoder-only models given the same training data~\cite {behnamghader2024llm2vec}.
We selected GPT-2 as the LLM to generate the prompt embeddings, which enhance the time series embeddings. The GPT-2 comprises a tokenizer and a GPT-2 model. All parameters in the GPT-2 are frozen. 
%The structure of the GPT-2 model is shown in Appendix.
%, and a padding layer.

\textbf{Pre-trained LLM.} The tokenizer is responsible for converting prompt input $\mathbf{P}_S \in \mathbb{R}^{S \times N}$ into a series of token IDs $\mathbf{P}_G \in \mathbb{R}^{G \times N}$, where $G$ represents the token ID number in a prompt. Subsequently, these prompt tokens are fed into the GPT-2 model to generate prompt embeddings:
\begin{align}
\overline{\mathcal{P}}_G^{i} &= \mathit{MMSA}(\mathit{LN}(\mathbf{P}^{i}_G)) + \mathbf{P}^{i}_G, \label{eq:mmsa_ln} \\
\mathcal{P}_{G}^{i+1} &= \mathit{FFN}(\mathit{LN}(\overline{\mathcal{P}}^{i}_G)) + \overline{\mathcal{P}}^{i}_G, \label{eq:ffn_ln} \\
%\mathit{LN}(\mathbf{P}^{i}_G) &= \gamma \odot \frac{\mathbf{P}^{i}_G - \mu}{\sigma} + \beta, \\
\mathit{MMSA}(\mathbf{P}^{i}_G) &= \phi_o (\mathit{Attention}(\phi_q \mathbf{P}^{i}_G, \phi_k \mathbf{P}^{i}_G, \phi_v \mathbf{P}^{i}_G)),
%\mathit{Attention}(\mathbf{P}^{i}_G) &= \mathit{softmax}\left(\frac{\mathbf{P}^{i}_G \mathbf{P}^{iT}_G}{\sqrt{d_k}}\right) \mathbf{P}^i_G, \\
%\mathit{FFN}(\overline{\mathcal{P}}^i_G) &= \max(0, \mathbf{W}_1\mathcal{P}^i_G + \mathbf{b}_1) \mathbf{W}_2 + \mathbf{b}_2,
\end{align}

\noindent where $\mathcal{\overline{P}}^i_G \in \mathbb{R}^{G \times N \times E}$ represents the intermediate representation of the $i_{\text{th}}$ layer after applying the $\mathit{MMSA}(\cdot)$ and the $\mathit{LN}(\cdot)$, $E$ denotes the hidden dimension of the GPT-2. 
$\mathbf{P}^{0}_G =\left[\mathbf{P}_G + \mathbf{PE}\right]$, $\mathbf{PE}$ represents the learnable positional encoding.
$\phi_o$, $\phi_q$, $\phi_k$, and $\phi_v$ are the linear projections. $\mathcal{P}^{i+1}_G \in \mathbb{R}^{G \times N \times E}$ symbolizes the output of GPT-2 .

\textbf{Last Token Embedding Storage}. %Recent studies have verified that 
It is verified that not all tokens are equally important for language model training~\cite{lin2024rho,behnamghader2024llm2vec}.
The last token in a prompt holds the most comprehensive knowledge due to the masked multi-self attention within the LLMs.
Specifically, the representation of the last token at position $G$ is influenced exclusively by the representations of its previous tokens at positions $\{1, 2, \cdots, G-1\}$. Thus, we tailor and store the well-trained last token embeddings $\mathbf{L}_N= \left\{\mathbf{l}_1, \ldots, \mathbf{l}_{N}\right\} \in \mathbb{R}^{N \times E}$ from the $\mathcal{P}^{i+1}_G$ to reduce computational costs.
%To maintain consistent sample embedding lengths under a variable, we repeatedly append the last token as padding until the number of tokens in the variable is the same for convenient storage.

%The padding ensures that inputs into the model maintain a consistent size, which is crucial for batch processing within neural networks. Collectively, these components allow the pre-trained LLM to enhance the dual-modality alignment by providing contextual embeddings that augment the Transformer's ability to interpret and forecast multivariate time series data.

\textbf{Prompt Encoder.} We define prompt encoder as $\mathit{PromptEncoder(\cdot)}$. Its structure follows the decoder in Pre-LN Transformer, identical to $\mathit{TSEncoder(\cdot)}$. We denote the output of $\mathit{PromptEncoder(\cdot)}$ as $\overline{\mathbf{L}}_N \in \mathbb{R}^{N \times E}$.
\if 0
\begin{equation}
\overline{\mathbf{L}}_N=\mathit{PrtEncoder}\left(\mathbf{W}_N\right),
\end{equation}

\noindent where the obtained variable tokens interact through self-attention and are independently processed by the shared feed-forward network within each encoder.
\fi
\subsection{Cross-Modality Alignment}
To aggregate the time series and the prompt modalities, we design a cross-modality alignment based on channel-wise similarity retrieval. It aims at using \textit{disentangled yet weak} time series embeddings $\overline{\mathbf{H}}_T \in \mathbb{R}^{C \times N}$ to retrieve \textit{disentangled and robust} time series embeddings $\overline{\mathbf{H}}_C \in \mathbb{R}^{N \times E}$ from \textit{entangled and robust} prompt embeddings $\overline{\mathbf{L}}_N \in \mathbb{R}^{C \times N}$.

First, we employ three linear layers $\psi_q, \psi_v, \psi_k$ to transform $\overline{\mathbf{H}}_T$ and $\overline{\mathbf{L}}_N$ to three compact embeddings: $\psi_q(\overline{\mathbf{H}}_T)$,~$\psi_v(\overline{\mathbf{L}}_N)$, and $\psi_k(\overline{\mathbf{L}}_N)$. Then, we compute the channel-wise similarity matrix $\mathbf{M}_T \in \mathbb{R}^{C \times E}$ by matrix multiplication followed by softmax:
\begin{equation}
\mathbf{M}_T=\mathit{F}_{\text{softmax}}\left(\psi_q\left(\overline{\mathbf{H}}_T\right) \otimes  \psi_k\left(\overline{\mathbf{L}}_N\right)\right),
\end{equation}

\noindent where $\otimes$ denotes matrix multiplication.

We perform channel-wise feature aggregation by restoring the channel dimension through the matrix multiplication of $\psi_v(\overline{\mathbf{L}}_N)$ with $\mathbf{M}_T$. Finally, we get the output by adding $\overline{\mathbf{H}}_T$ to it by matrix addition:
\begin{equation}
    \overline{\mathbf{H}}_C = \omega^c\left(\psi_v\left(\overline{\mathbf{L}}_N\right) \otimes  \mathbf{M}_T\right) \oplus \overline{\mathbf{H}}_T,
\end{equation}

\noindent where
% $\overline{\mathbf{H}}_C \in \mathbb{R}^{C \times N}$. 
$\omega^c$ is the linear layer and $\oplus$ denotes addition.

Through cross-modality alignment, we transfer the knowledge learned from the pre-trained LLM into time series embeddings, which thus improves the model performance.

%\begin{table}[htbp]
%\caption{Statistics of time series and prompt.}
%\begin{tabular}{cccccccc}
%\toprule
%Dataset      & Domain      & Dim & Dataset Size          & Frequency & Instance & Prompt Token & Split \\ \midrule
%ETTm1 & Temperature & 7   & (34465, 11521, 11521) & 15 min    & 57,600 & 483          & 6:2:2 \\
%ETTm2 & Temperature & 7   & (34465, 11521, 11521) & 15 min    & 57,600 & 493          & 6:2:2 \\
%ETTh1 & Temperature & 7   & (8545, 2881, 2881)    & 1 hour    & 14,400 & 444          & 6:2:2 \\
%ETTh2 & Temperature & 7   & (8545, 2881, 2881)    & 1 hour    & 14,400 & 505          & 6:2:2 \\
%ECL          & Electricity & 321 & (18317, 2633, 5261)   & 1 hour    & 26,304 & 369          & 7:1:2 \\
%FRED-MD         & Economic    & 107 & (450, 51,122)         & 1 month   & 728    & 230          & 7:1:2 \\
%ILI          & Health      & 7   & (617, 74, 170)        & 1 week    & 966    & 232          & 7:1:2 \\
%Weather      & Environment & 21  & (36792, 5271, 10540)  & 10 min    & 52,696 & 508          & 7:1:2 \\ \bottomrule
%\end{tabular}
%\label{tab:data}
%\end{table}

\subsection{Time Series Forecasting}
We design a time series forecasting module including a multivariate  
Transformer decoder and a projection function. In particular, we input the aligned time series embeddings $\overline{\mathbf{H}}_C$ into the multivariate Transformer decoder $\mathit{MTDecoder(\cdot)}$ to map the dependencies among variables. Finally, we use a projection function for final forecasting.

% The $\mathit{MTDecoder(\cdot)}$ is a lightweight Pre-LN Transformer decoder. 
We first feed the $\overline{\mathbf{H}}_C$ into a layer normalization layer $LN(\cdot)$ to obtain normalized embeddings $\widetilde{\mathbf{H}}_C^{i}$. Then, we employ a masked multi-self attention layer $MMSA(\cdot)$ with residual connection to obtain $\overline{\mathbf{H}}_C^{i}$.

Then, $\overline{\mathbf{H}}_C^{i}$ is fed to the second layer normalization $LN(\cdot)$ followed by a multi-head cross-attention layer $MHCA(\cdot)$:
\begin{equation}
    \mathbf{\check{H}}_C^{i} = \mathit{MHCA}(\mathit{LN}(\overline{\mathbf{H}}_C^{i})) + \mathbf{\overline{H}}^{i}_C,
\end{equation}
\begin{equation}
    \mathit{MHCA}(\mathbf{H}^{i}_C) = \varsigma_o (\mathit{Attention}(\varsigma _q \mathbf{\overline{H}}^{i}_C, \varsigma_k \mathbf{\overline{H}}^{i}_C, \varsigma_v \mathbf{\overline{H}}^{i}_C)),
\end{equation}
where $\varsigma_o$, $\varsigma_q$, $\varsigma_k$, and $\varsigma_v$ are linear projections. We apply residual connection to obtain the output $\mathbf{\check{H}}_C$ of $\mathit{MTDecoder(\cdot)}$.

Finally, the $\mathbf{\check{H}}_C$ is input into a projection function for future prediction, which is formulated as follows:
%The processed outputs are then channeled into a linear projection module designed to forecast future values in the time series:
\begin{equation}
    \widehat{\mathbf{X}}_M = \mathbf{W}_p \mathbf{\check{H}}_C + \mathbf{b}_p,
\end{equation}
\noindent where $\widehat{\mathbf{X}}_M \in \mathbb{R}^{M \times N}$ denotes the projected embeddings.
%$\mathbf{W}_p$ and $\mathbf{b}_p$ are the learnable parameters. 
Finally, we denormalize the $\widehat{\mathbf{X}}_M$.

\subsection{Overall Objective Function}
%To optimize the forecasting performance, we employ the mean squared error (MSE) as our loss function, combined with L2 regularization to mitigate overfitting:
%\begin{equation}
%%\mathcal{L} = \left\| \mathbf{\widehat{X}}_M - \mathbf{X}_M \right\|^2 + \lambda \cdot L_{\text{reg}},
%\label{eq:loss}
%\end{equation}
%\noindent where $\mathbf{\widehat{X}}_M$ represents the predicted time series, $\mathbf{X}_M$ is the ground truth, and $L_{\text{reg}}$ is the L2 regularization term. The coefficient $\lambda$ is a hyperparameter that balances the regularization strength relative to the MSE.
The loss function of TimeCMA contains two parts: a prediction loss $L_{pre}$ and a regularization loss $L_{reg}$. We combine them and the overall loss is as follows, 
\begin{equation}
    L_{task} = L_{pre} + \lambda L_{reg},
\end{equation}
where $\lambda$ is a weight to trade off the prediction and regularization losses. We use Mean Squared Error as the prediction loss, \re{i.e.,}
%\begin{equation}
$L_{pre} = \frac{1}{\mathcal{M}}\sum_{M=1}^{\mathcal{M}} (\mathbf{\widehat{X}}_M - \mathbf{X}_M)^2$,
%\end{equation}
where $\mathcal{M}$ is the training sample size, and $L_{reg}$ is $L_2$ regularization. 
%In addition, we use L2 regularization ~\cite{DBLP:books/lib/HastieTF09} as the regularization loss $L_{reg}$.

\begin{table*}[ht]
\centering
% \vspace{-0.5cm}
% \setlength{\tabcolsep}{1.5pt}
% \tiny
\resizebox{0.9\textwidth}{!}{
\begin{tabular}{cc|cc|cc|cc|cc|cc|cc|cc|cc}
\toprule
\multicolumn{2}{c|}{Models}                         & \multicolumn{2}{c|}{TimeCMA}                      & \multicolumn{2}{c|}{Time-LLM}                      & \multicolumn{2}{c|}{UniTime}                      & \multicolumn{2}{c|}{OFA}                           & \multicolumn{2}{c|}{iTransformer}                 & \multicolumn{2}{c|}{PatchTST}                     & \multicolumn{2}{c|}{TimesNet}                     & \multicolumn{2}{c}{Dlinear}              \\ \midrule
\multicolumn{2}{c|}{Metric}                         & MSE                     & MAE                     & MSE                      & MAE                     & MSE                     & MAE                     & MSE                      & MAE                     & MSE                     & MAE                     & MSE                     & MAE                     & MSE                     & MAE                     & MSE            & MAE                     \\ \midrule
\multicolumn{1}{c|}{\multirow{5}{*}{ETTm1}}   & 96  & \textbf{0.312}          & \textbf{0.351}          & 0.359                    & 0.381                   & \ul{0.322} & \ul{0.363} & 0.335                    & 0.369                   & 0.334                   & 0.368                   & 0.344                   & 0.373                   & 0.338                   & 0.375                   & 0.345          & 0.372                   \\
\multicolumn{1}{c|}{}                         & 192 & \textbf{0.361}          & \textbf{0.378}          & 0.383                    & 0.393                   & \ul{0.366} & 0.387                   & 0.374                    & \ul{0.385} & 0.377                   & 0.391                   & 0.367                   & 0.386                   & 0.374                   & 0.387                   & 0.380          & 0.389                   \\
\multicolumn{1}{c|}{}                         & 336 & \textbf{0.392}          & \textbf{0.401}          & 0.416                    & 0.414                   & 0.398                   & \ul{0.407} & 0.407                    & 0.406                   & 0.426                   & 0.420                   & \textbf{0.392}          & 0.407                   & 0.410                   & 0.411                   & 0.413          & 0.413                   \\
\multicolumn{1}{c|}{}                         & 720 & \ul{0.453} & \textbf{0.438}          & 0.483                    & 0.449                   & \textbf{0.454}          & \ul{0.440} & \ul{0.469}  & 0.442                   & 0.491                   & 0.459                   & 0.464                   & 0.442                   & 0.478                   & 0.450                   & 0.474          & 0.453                   \\
\multicolumn{1}{c|}{}                         & Avg & \textbf{0.380}          & \textbf{0.392}          & 0.410                    & 0.409                   & \ul{0.385} & \ul{0.399} & 0.396                    & 0.401                   & 0.407                   & 0.410                   & 0.392                   & 0.402                   & 0.400                   & 0.406                   & 0.403          & 0.407                   \\ \midrule
\multicolumn{1}{c|}{\multirow{5}{*}{ETTm2}}   & 96  & \textbf{0.173}          & \textbf{0.258}          & 0.193                    & 0.280                   & 0.183                   & 0.266                   & 0.190                    & 0.275                   & 0.180                   & 0.264                   & \ul{0.177} & \ul{0.260} & 0.187                   & 0.267                   & 0.193          & 0.292                   \\
\multicolumn{1}{c|}{}                         & 192 & \textbf{0.238}          & \textbf{0.301}          & 0.257                    & 0.318                   & 0.251                   & 0.310                   & 0.253                    & 0.313                   & 0.250                   & 0.309                   & \ul{0.246} & \ul{0.305} & 0.249                   & 0.309                   & 0.284          & 0.362                   \\
\multicolumn{1}{c|}{}                         & 336 & \textbf{0.297}          & \textbf{0.338}          & 0.317                    & 0.353                   & 0.319                   & 0.351                   & 0.321                    & 0.360                   & 0.311                   & 0.348                   & \ul{0.305} & \ul{0.343} & 0.321                   & 0.351                   & 0.369          & 0.427                   \\
\multicolumn{1}{c|}{}                         & 720 & \textbf{0.393}          & \textbf{0.394}          & 0.419                    & 0.411                   & 0.420                   & 0.410                   & 0.411                    & 0.406                   & 0.412                   & 0.407                   & 0.410                   & 0.405                   & \ul{0.408} & \ul{0.403} & 0.554          & 0.522                   \\
\multicolumn{1}{c|}{}                         & Avg & \textbf{0.275}          & \textbf{0.323}          & 0.296                    & 0.340                   & 0.293                   & 0.334                   & 0.294                    & 0.339                   & 0.288                   & 0.332                   & \ul{0.285} & \ul{0.328} & 0.291                   & 0.333                   & 0.350          & 0.401                   \\ \midrule
\multicolumn{1}{c|}{\multirow{5}{*}{ETTh1}}   & 96  & \textbf{0.373}          & \textbf{0.391}          & 0.398                    & 0.410                   & 0.397                   & 0.418                   & 0.398                    & 0.424                   & 0.386                   & 0.405                   & 0.404                   & 0.413                   & \ul{0.384}             & 0.402                   & 0.386          & \ul{0.400} \\
\multicolumn{1}{c|}{}                         & 192 & \textbf{0.427}          & \textbf{0.421}          & 0.451                    & 0.440                   & \ul{0.434}             & 0.439                   & 0.449                    & \ul{0.427} & 0.441                   & 0.436                   & 0.454                   & 0.430                   & \ul{0.434}             & 0.429                   & 0.437          & 0.432                   \\
\multicolumn{1}{c|}{}                         & 336 & \textbf{0.458}          & \textbf{0.448}          & 0.473                    & \ul{0.451} & \ul{0.468}             & 0.457                   & 0.492                    & 0.466                   & 0.487                   & 0.458                   & 0.497                   & 0.462                   & 0.491                   & 0.469                   & 0.481          & 0.459                   \\
\multicolumn{1}{c|}{}                         & 720 & \textbf{0.449}          & \textbf{0.460}          & \ul{0.469}  & \ul{0.470} & 0.469                   & 0.477                   & 0.487                    & 0.483                   & 0.503                   & 0.491                   & 0.496                   & 0.481                   & 0.521                   & 0.500                   & 0.519          & 0.516                   \\
\multicolumn{1}{c|}{}                         & Avg & \textbf{0.423}          & \textbf{0.431}          & 0.448                    & \ul{0.443} & \ul{0.442} & 0.448                   & 0.457                    & 0.450                   & 0.454                   & 0.447                   & 0.463                   & 0.449                   & 0.458                   & 0.450                   & 0.456          & 0.452                   \\ \midrule
\multicolumn{1}{c|}{\multirow{5}{*}{ETTh2}}   & 96  & \textbf{0.286}          & \textbf{0.336}          & \ul{0.295}  & \ul{0.345} & 0.296                   & \ul{0.345} & 0.312                    & 0.360                   & 0.297                   & 0.349                   & 0.312                   & 0.358                   & 0.340                   & 0.374                   & 0.333          & 0.387                   \\
\multicolumn{1}{c|}{}                         & 192 & \textbf{0.363}          & \textbf{0.387}          & 0.386                    & 0.399                   & \ul{0.374} & \ul{0.394} & 0.387                    & 0.405                   & 0.380                   & 0.400                   & 0.397                   & 0.408                   & 0.402                   & 0.414                   & 0.477          & 0.476                   \\
\multicolumn{1}{c|}{}                         & 336 & \textbf{0.406}          & \textbf{0.421}          & 0.419                    & 0.429                   & \ul{0.415} & \ul{0.427} & 0.424                    & 0.437                   & 0.428                   & 0.432                   & 0.435                   & 0.440                   & 0.452                   & 0.452                   & 0.594          & 0.541                   \\
\multicolumn{1}{c|}{}                         & 720 & \textbf{0.417}          & \textbf{0.438}          & \ul{0.425}  & \ul{0.442} & \ul{0.425} & 0.444                   & 0.433                    & 0.453                   & 0.427                   & 0.445                   & 0.436                   & 0.449                   & 0.462                   & 0.468                   & 0.831          & 0.657                   \\
\multicolumn{1}{c|}{}                         & Avg & \textbf{0.372}          & \textbf{0.397}          & 0.381                    & 0.404                   & \ul{0.378} & \ul{0.403} & 0.389                    & 0.414                   & 0.383                   & 0.407                   & 0.395                   & 0.414                   & 0.414                   & 0.427                   & 0.559          & 0.515                   \\ \midrule
\multicolumn{1}{c|}{\multirow{5}{*}{ECL}}     & 96  & \textbf{0.143}          & \textbf{0.238}          & 0.172                    & 0.265                   & 0.196                   & 0.287                   & 0.197                    & 0.290                   & \ul{0.148} & \ul{0.240} & 0.186                   & 0.269                   & 0.168                   & 0.272                   & 0.197          & 0.282                   \\
\multicolumn{1}{c|}{}                         & 192 & \textbf{0.161}          & \ul{0.259} & 0.182                    & 0.279                   & 0.199                   & 0.291                   & 0.201                    & 0.292                   & \ul{0.162} & \textbf{0.253}          & 0.190                   & 0.273                   & 0.184                   & 0.289                   & 0.196          & 0.285                   \\
\multicolumn{1}{c|}{}                         & 336 & \textbf{0.169}          & \textbf{0.261}          & 0.195                    & 0.288                   & 0.214                   & 0.305                   & 0.217                    & 0.309                   & \ul{0.178} & \ul{0.269} & 0.206                   & 0.290                   & 0.198                   & 0.300                   & 0.209          & 0.301                   \\
\multicolumn{1}{c|}{}                         & 720 & \textbf{0.219}          & \textbf{0.315}          & 0.233                    & 0.320                   & 0.254                   & 0.335                   & 0.253                    & 0.339                   & 0.225                   & \ul{0.317} & 0.247                   & 0.322                   & \ul{0.220} & 0.320                   & 0.245          & 0.333                   \\
\multicolumn{1}{c|}{}                         & Avg & \textbf{0.174}          & \textbf{0.269}          & 0.195                    & 0.288                   & 0.216                   & 0.306                   & 0.217                    & 0.308                   & \ul{0.178} & \ul{0.270} & 0.207                   & 0.289                   & 0.192                   & 0.295                   & 0.212          & 0.300                   \\ \midrule
\multicolumn{1}{c|}{\multirow{5}{*}{FRED}}    & 24  & \textbf{22.702}         & \textbf{0.864}          & \ul{27.285} & \ul{0.875} & 31.178                  & 0.931                   & 28.317                   & 0.947                   & 28.017                  & 0.893                   & 35.777                  & 1.014                   & 43.268                  & 1.266                   & 37.898         & 1.070                   \\
\multicolumn{1}{c|}{}                         & 36  & \textbf{40.880}         & \textbf{1.157}          & \ul{48.730} & \ul{1.172} & 54.172                  & 1.223                   & 59.520                   & 1.306                   & 50.837                  & 1.274                   & 61.034                  & 1.345                   & 69.514                  & 1.533                   & 71.047         & 1.477                   \\
\multicolumn{1}{c|}{}                         & 48  & \textbf{60.045}         & \textbf{1.352}          & \ul{73.494} & \ul{1.460} & 83.836                  & 1.518                   & 74.808                   & 1.516                   & 78.018                  & 1.793                   & 93.482                  & 1.667                   & 89.913                  & 1.742                   & 118.579        & 2.002                   \\
\multicolumn{1}{c|}{}                         & 60  & \textbf{65.015}         & \textbf{1.509}          & 108.221                  & 1.758                   & 118.429                 & 1.830                   & \ul{83.613} & \ul{1.641} & 90.212                  & 1.693                   & 133.444                 & 2.011                   & 116.187                 & 1.976                   & 156.844        & 2.221                   \\
\multicolumn{1}{c|}{}                         & Avg & \textbf{48.161}         & \textbf{1.221}          & 64.433                   & \ul{1.316} & 71.901                  & 1.376                   & \ul{61.565} & 1.353                   & 61.771                  & 1.413                   & 80.934                  & 1.509                   & 79.721                  & 1.629                   & 96.092         & 1.693                   \\ \midrule
\multicolumn{1}{c|}{\multirow{5}{*}{ILI}}     & 24  & \textbf{1.996}          & 0.998                   & 2.383                    & 1.004                   & 2.346                   & 0.954                   & 2.732                    & 1.100                   & 2.347                   & 1.731                   & 2.335                   & \ul{0.989} & \ul{2.317} & \textbf{0.934}          & 2.398          & 1.040                   \\
\multicolumn{1}{c|}{}                         & 36  & \textbf{1.906}          & \textbf{0.915}          & 2.390                    & 0.993                   & 1.998                   & \ul{0.912} & 2.664                    & 1.063                   & 2.468                   & 0.998                   & 2.561                   & 1.035                   & \ul{1.972} & 0.920                   & 2.646          & 1.088                   \\
\multicolumn{1}{c|}{}                         & 48  & \textbf{1.867}          & \textbf{0.868}          & 2.394                    & 1.003                   & \ul{1.979} & \ul{0.912} & 2.617                    & 1.041                   & 2.489                   & 1.016                   & 2.465                   & 1.022                   & 2.238                   & 0.913                   & 2.614          & 1.086                   \\
\multicolumn{1}{c|}{}                         & 60  & \textbf{1.920}          & \textbf{0.904}          & 2.562                    & 1.049                   & 2.109                   & 0.938                   & 2.478                    & 1.035                   & 2.471                   & 1.065                   & 2.189                   & 0.997                   & \ul{2.027} & \ul{0.928} & 2.804          & 1.146                   \\
\multicolumn{1}{c|}{}                         & Avg & \textbf{1.922}          & \textbf{0.921}          & 2.432                    & 1.012                   & \ul{2.108} & \ul{0.929} & 2.623                    & 1.060                   & 2.444                   & 1.203                   & 2.388                   & 1.011                   & 2.139                   & 0.931                   & 2.616          & 1.090                   \\ \midrule
\multicolumn{1}{c|}{\multirow{5}{*}{Weather}} & 96  & \textbf{0.167}          & \textbf{0.211}          & 0.198                    & 0.235                   & 0.171                   & \ul{0.214} & 0.203                    & 0.244                   & 0.174                   & \ul{0.214} & 0.177                   & 0.218                   & 0.172                   & 0.220                   & 0.196          & 0.255                   \\
\multicolumn{1}{c|}{}                         & 192 & \textbf{0.212}          & \textbf{0.253}          & 0.240                    & 0.269                   & \ul{0.217} & \ul{0.254} & 0.247                    & 0.277                   & 0.221                   & \ul{0.254} & 0.222                   & 0.259                   & 0.219                   & 0.261                   & 0.237          & 0.296                   \\
\multicolumn{1}{c|}{}                         & 336 & \textbf{0.270}          & \textbf{0.292}          & 0.295                    & 0.308                   & \ul{0.274} & \ul{0.293} & 0.297                    & 0.311                   & 0.278                   & 0.296                   & 0.277                   & 0.297                   & 0.280                   & 0.306                   & 0.283          & 0.335                   \\
\multicolumn{1}{c|}{}                         & 720 & \ul{0.350} & \textbf{0.348}          & 0.368                    & 0.353                   & 0.351                   & 0.343                   & 0.368                    & 0.356                   & 0.358                   & \ul{0.349} & 0.352                   & 0.347                   & 0.365                   & 0.359                   & \textbf{0.345} & 0.381                   \\
\multicolumn{1}{c|}{}                         & Avg & \textbf{0.250}          & \textbf{0.276}          & 0.275                    & 0.291                   & \ul{0.253} & \textbf{0.276}          & 0.279                    & 0.297                   & 0.258                   & \ul{0.278} & 0.257                   & 0.280                   & 0.259                   & 0.287                   & 0.265          & 0.317                   \\
\bottomrule
\end{tabular}}
\caption{\small Forecasting performance comparisons. The input sequence length is 36 for the Illness and FRED datasets and 96 for others. 
% Predictive lengths vary based on the dataset, set to \{24, 36, 48, 60\} for Illness and FRED datasets, and \{96, 192, 336, 720\} for others. 
%Note that we bold the best performance and underline the second-best performance.
}
% \vspace{-10pt}
\label{tab:main}
\end{table*}

\section{Experiments}
\label{experiment}
% \subsection{Experimental Setup}

\subsubsection{Datasets.} We conduct experiments on eight datasets: ETTm1, ETTm2, ETTh1, ETTh2~\cite{Zeng2022AreTE}, ECL~\cite{asuncion2007uci}, FRED-MD~\cite{mccracken2016fred}, ILI and Weather~\cite{DBLP:conf/nips/WuXWL21}. 
We removed variables with missing values in the FRED-MD~\cite{DBLP:journals/corr/abs-2403-20150} and simplified it as FRED. 
% Details of time series and prompt datasets are provided in Appendix.

\subsubsection{Baselines and Evaluation.} 
We evaluate seven baseline models across five categories:
(1) Prompt-based LLMs: Time-LLM~\cite{jin2023time}, UniTime~\cite{liu2024unitime}. 
(2)Time series-based LLM: OFA~\cite{DBLP:conf/nips/ZhouNW0023}.
(3) Transformer-based models: iTransformer~\cite{DBLP:journals/corr/abs-2310-06625}, and PatchTST~\cite{Yuqietal-2023-PatchTST}.
% , and FEDformer~\cite{DBLP:conf/icml/ZhouMWW0022}. 
(4) Linear-based method: Dlinear~\cite{Zeng2022AreTE}.
(5) CNN-based method: TimesNet~\cite{wu2023timesnet}.
The evaluation metrics are mean square error (MSE) and mean absolute error (MAE). 
The test batch size is set to 1 for all methods to guarantee fairness during testing. Each experiment is repeated at least three times with different seeds on NVIDIA A100 GPUs. 
% More implementation details in Appx.%Refer to the Appendix of Implementation Details.

%\subsubsection{Implementation Details.} To ensure fairness and conserve general computational cost for MTSF, all LLM-based models use GPT-2 as the backbone. Most importantly, the test batch size is set to 1 for all models to guarantee consistency in the testing phase~\cite{DBLP:journals/corr/abs-2403-20150}. The lookback window length is fixed at 36 for the ILI and FRED, and 96 for the others. We train our model using the AdamW optimizer and select the trained model with the lowest average validation loss for testing. Each experiment is repeated at least three times with different seeds on NVIDIA A100 GPUs.

\subsubsection{Main Results.}
% \label{sec:main_result}
Table~\ref{tab:main} illustrates the average performance of TimeCMA outperforms all baselines in all cases. (1) \emph{LLM-based models perform better than deep learning and linear models.} These results verify our motivation to use LLMs for multivariate time series forecasting. (2) \emph{Inverted embedding is essential for capturing multivariate dependencies.} For datasets with more variables, TimeCMA can perform better since we introduce inverted embedding and multivariate attention into the TimeCMA.
(3) \emph{Prompt-based LLMs outperform time series-based LLMs.} The prompt-based LLM, such as TimeCMA, outperforms the time series-based LLM, OFA, with an average improvement of 16.1\% in MSE and 11.9\% in MAE. This indicates that the prompt enhanced the time series embeddings. Compared to UniTime, TimeCMA shows an average improvement of about 13.9\% in MSE and 12.6\% in MAE. 

\begin{figure}[b]    
\includegraphics[width=\columnwidth]{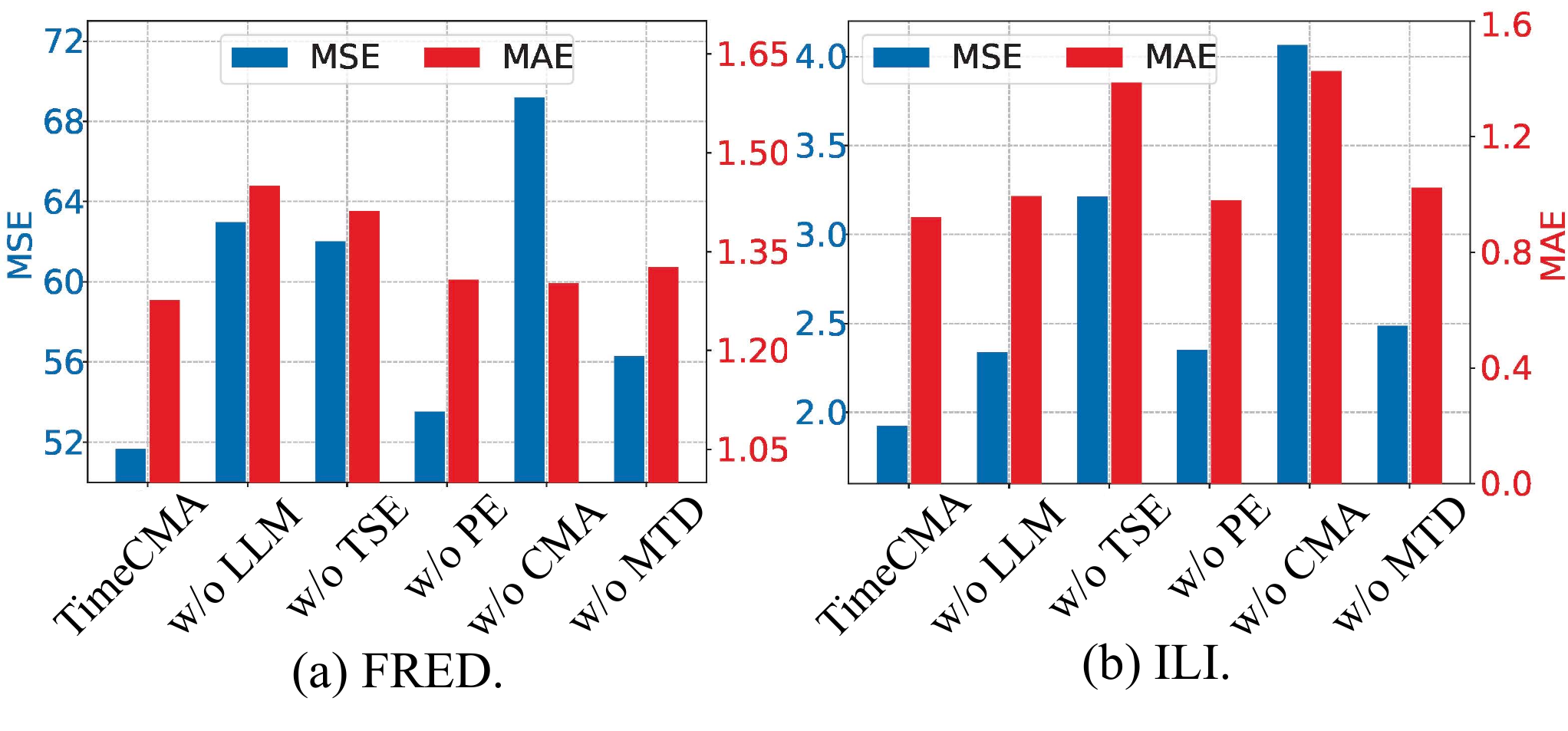}   
\caption{Ablation study of model design.}
\label{fig:ab_model}
\end{figure}

\subsubsection{Ablation Studies of Model Design.}
% To better understand the model designs in TimeCMA, we constructed variants of TimeCMA and conducted ablation studies. The experimental results, which are average values across all predictive lengths, are summarized in Fig.~\ref{fig:ab_model}. 
% The structures of these designed models are detailed in Appendix. 
% We also conducted ablation studies with different prompts, which are reported in Appendix Figures A2 and A3. 

Fig.~\ref{fig:ab_model} indicates the ablation studies of model design, which are average values across all predictive lengths.
The variant with the most significant impact is cross-modality alignment (\textit{w/o CMA}), where CMA is replaced with concatenation. The results highlight that our similarity-based retrieval of cross-modal design is superior to simple concatenation.
The next most impactful variant is the LLM. The result for \textit{w/o LLM} signifies the LLM-empowered dual branches have better prediction results than the time series branch.
Without a time series encoder (\textit{w/o~TSE}), the degradation results indicate that extracting disentangled time series embeddings is fundamental for forecasting. 
We find that removing the prompt encoder (\textit{w/o~PE}) has the least impact, as the LLM captures the dependencies between variables, and the prompt encoder's role is to prepare for the subsequent cross-modality alignment.
Furthermore, without multivariate Transformer decoder (\textit{w/o~MTD}) shows that decoding long-term temporal dependencies between multiple variables is essential for MTSF.

% \begin{figure}[t]
% \centering
%     \begin{minipage}[b]{0.495\linewidth}
%         \centering
%         \includegraphics[width=\linewidth]{figures/Ablation_FRED.pdf}
%         \caption*{(a) FRED.}
%     \end{minipage}
%     \hfill
%     \begin{minipage}[b]{0.495\linewidth}
%         \centering
%         \includegraphics[width=\linewidth]{figures/Ablation_ILI.pdf}
%         \caption*{(b) ILI.} 
%     \end{minipage}
%     \caption{Ablation study of model design.} 
%     \label{fig:ab_model}
% \end{figure}

\begin{figure}[t]    
\includegraphics[width=\columnwidth]{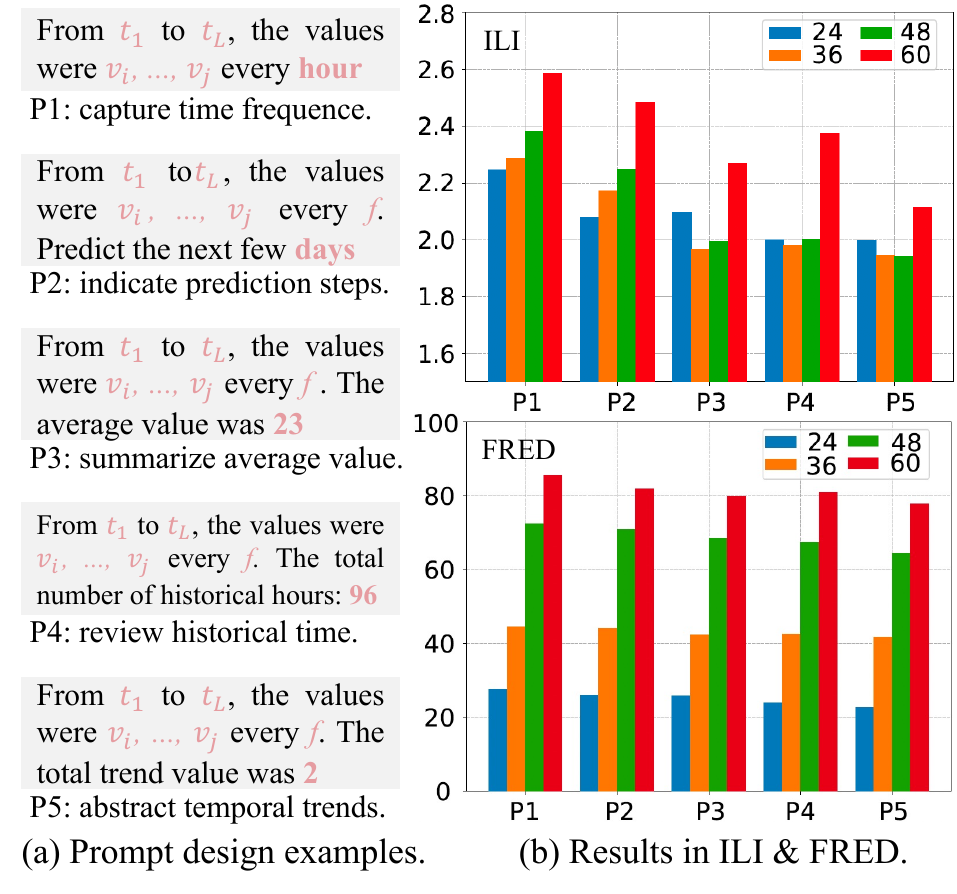}  
% \vspace{-10pt}
\caption{\small Five prompts with different purposes to trigger last token.}
% , time-series related token in pink: numerical-ended prompts (3,4,5) are better; prompt 5 abstracting time series trends is the best.}
\label{fig:ab_Prompt}
% \vspace{-10pt}
\end{figure}

\begin{table}[t]
\centering
\resizebox{0.9\columnwidth}{!}{
\begin{tabular}{c|ccc|ccc}
\toprule
Dataset  & \multicolumn{3}{c|}{ETTm1~-~96}                     & \multicolumn{3}{c}{ETTm2~-~96}                      \\ \midrule
Metric   & Param.         & Mem.         & Speed          & Param.         & Mem.         & Speed          \\ \midrule
Time-LLM & 44.66         & 28,882        & 1.08          & 44.95         & 29,140        & 1.08          \\
UniTime  & 108.54        & 4,168        & 0.39          & 108.54        & 4,168        & 0.39          \\
OFA      & \textbf{1.75}          & \ul{914}         & \ul{0.18}          & \textbf{1.74}          & \ul{914}         & \ul{0.17}          \\
TimeCMA  &  \ul{17.99}   & \textbf{821}    & \textbf{0.09}    & \ul{17.99}   & \textbf{818}    & \textbf{0.08}    \\ \bottomrule
\end{tabular}}
\caption{\small Efficiency analysis of LLM-based baselines.}
% , with Param. in (M), Mem. in (MiB), and Inference Speed. in(s/iter)}
\label{tab:efficiency}
\end{table}

% \begin{figure}[t]    
% \includegraphics[width=\columnwidth]{figures/prompt_design_and_result_big.pdf}   
% % \vspace{-10pt}
% \caption{\small \re{Five prompts with different purposes to trigger the last token, time-series related token in pink: numerical-ended prompts (3,4,5) are better; prompt 5 abstracting time series trends is the best.}}
% \label{fig:ab_Prompt}
% % \vspace{-10pt}
% \end{figure}

\subsubsection{Ablation Studies of Prompt Design.} We design five prompts: Prompts 1 to 5 are in Fig.~\ref{fig:ab_Prompt} (a), \re{with 
 different intentions for the LLMs on the last token, e.g. from ``to capture the frequency'' to ``summarize the trend''}. %, and Prompt 5 is in the lower left corner of Fig.~\ref{fig:framework}. 
%First, we extract time and value information directly in Prompt 1. Since we use the last token in the prompt to assist in time series forecasting, we carefully design it so that its last token is a numerical value to contain more time-related information. Prompt 2 contains the forecasting guidance for the next few timestamps. We summarize the average value of a variable over a given period in Prompt 3. Historical time information is derived in Prompt 4. We summarize the trend value in Prompt 5. 
The ablation studies of prompt design are demonstrated in Fig.~\ref{fig:ab_Prompt} (b) on MSE. A key insight is: \emph{prompts where the last token is a numerical value generally have better prediction performance}, such as Prompts 3, 4, and 5. Among these numeric last-token prompts, Prompt 5 is the best since it abstracts the time series trends. The second best is prompt 3, which averages the time series but may introduce noise since the average information is not necessarily useful for forecasting. Following this is Prompt 2, which emphasizes the historical time information. %Compared with time, values are more important for forecasting.

% \subsection{Model Analysis}

% \vspace{-0.1cm}
\subsubsection{Model Efficiency Analysis.}
Table~\ref{tab:efficiency} provides an efficiency analysis of TimeCMA, Time-LLM, and OFA. UniTime cannot be fairly compared in terms of efficiency because it is trained on all datasets.
% OFA uses six-layer LLMs, while TimeCMA use twelve layers. 
To ensure fairness of memory, we set the training batch size to 8, thus each iteration has 8 samples. The results show that TimeCMA has smaller training parameters and memory usage thanks to our design of the last token only and its storage. Conversely, UniTime has the largest parameters and Time-LLM has the largest memory usage and slowest speed. OFA's memory usage and inference speed are second only to TimeCMA, even though it only uses time series as the input. This shows that the designed prompt does not increase computational costs and essentially improves the prediction.% effect of the TimeCMA.

\begin{figure}[t]
%\vspace{-0.2cm}
\includegraphics[width=\columnwidth]{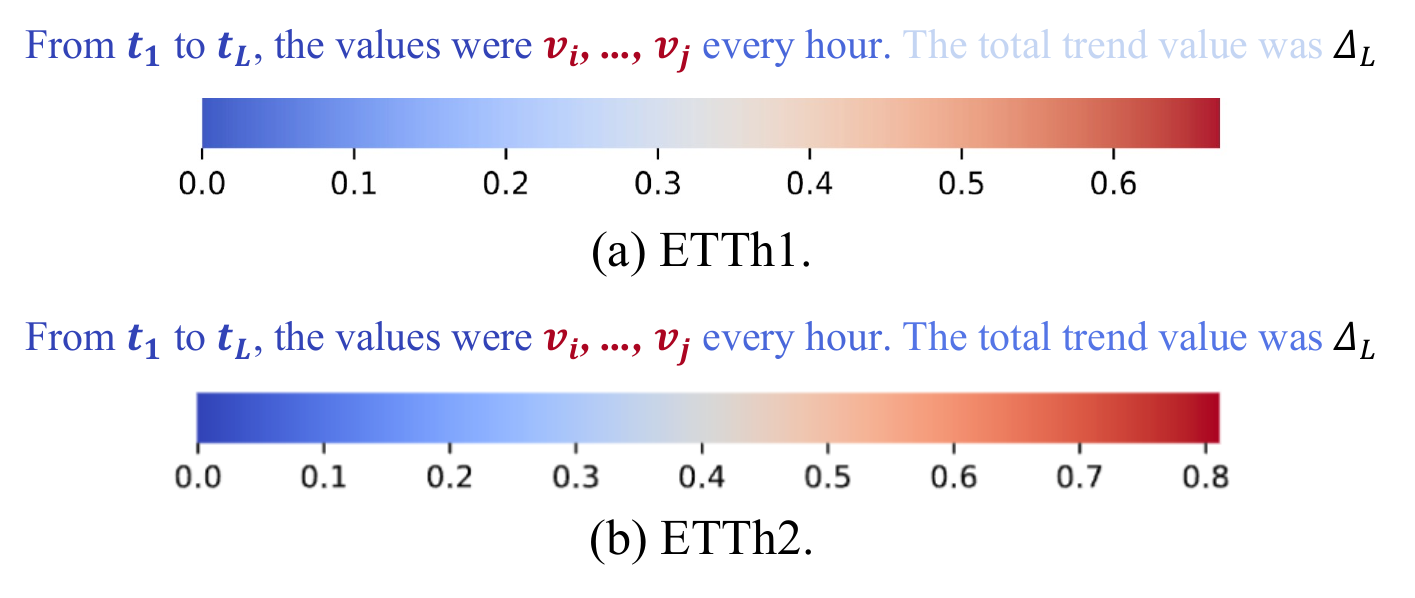} 
% \vspace{-20pt}
    \caption{Last token attention visualization}
    % : last token effectively encapsulate temporal information in the prompt.}
    \label{fig:last_token}
    % \vspace{-10pt}
\end{figure}

\begin{figure}[t]
\includegraphics[width=\columnwidth]{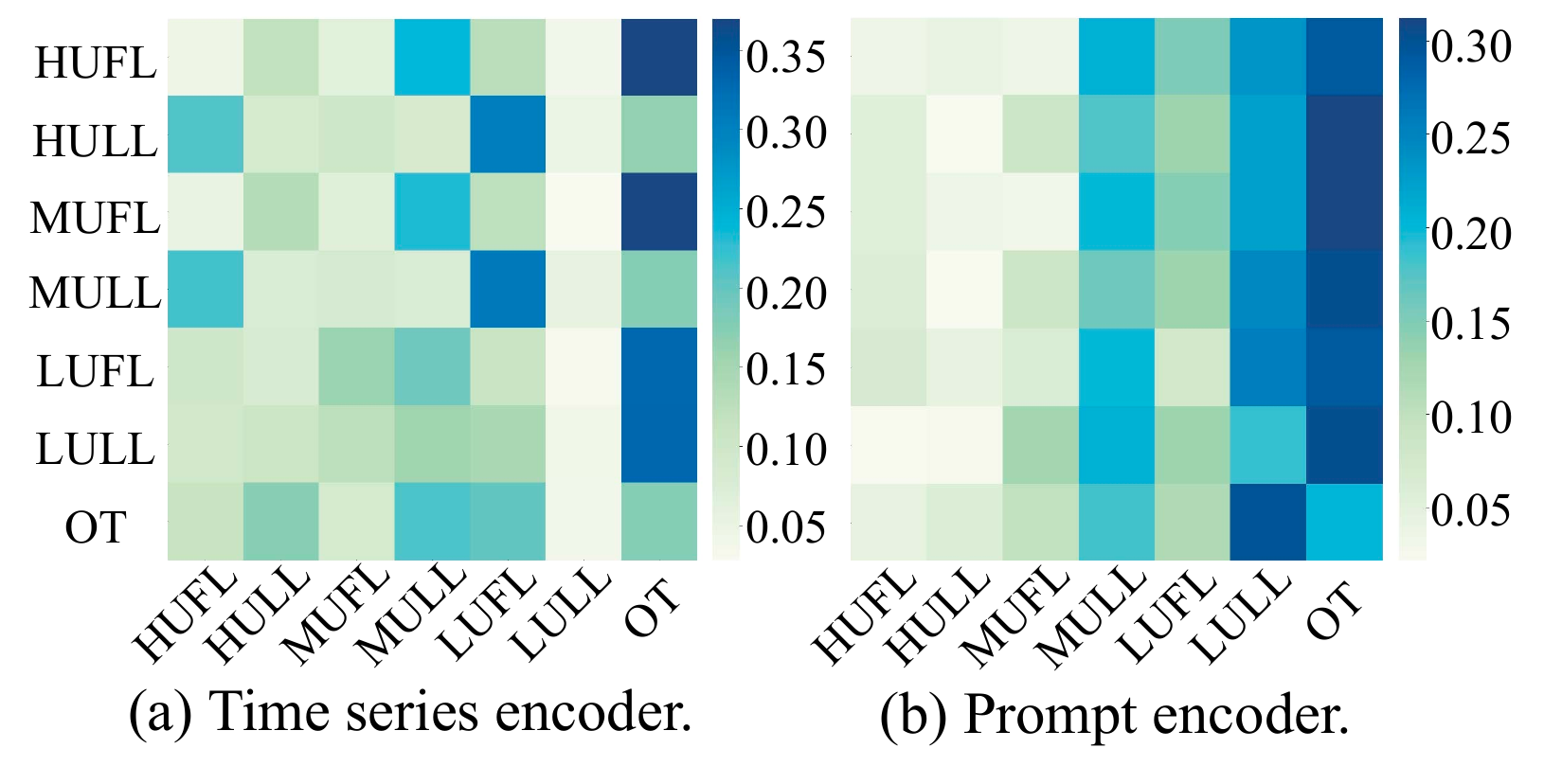} 
    \caption{\small Attention maps of Transformer and LLM encoders.}
    \label{fig:encoder_att_vis}
\end{figure}

\if 0
\begin{figure}[t]
\centering
    \begin{minipage}[b]{0.495\linewidth}
        \centering
        \includegraphics[width=\linewidth]{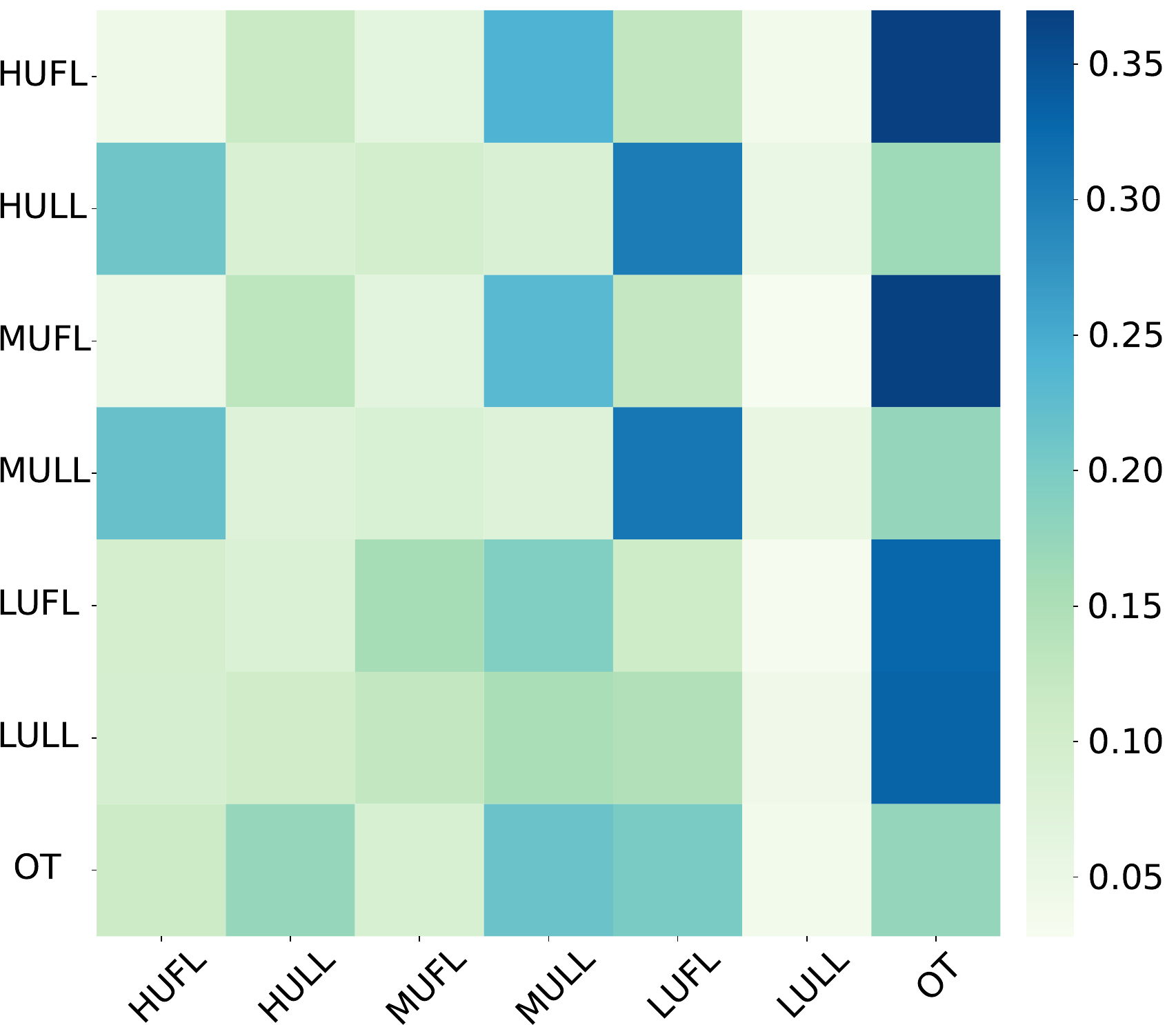}
        \caption*{(a) Time series encoder.}
    \end{minipage}
    \hfill
    \begin{minipage}[b]{0.495\linewidth}
        \centering
        \includegraphics[width=\linewidth]{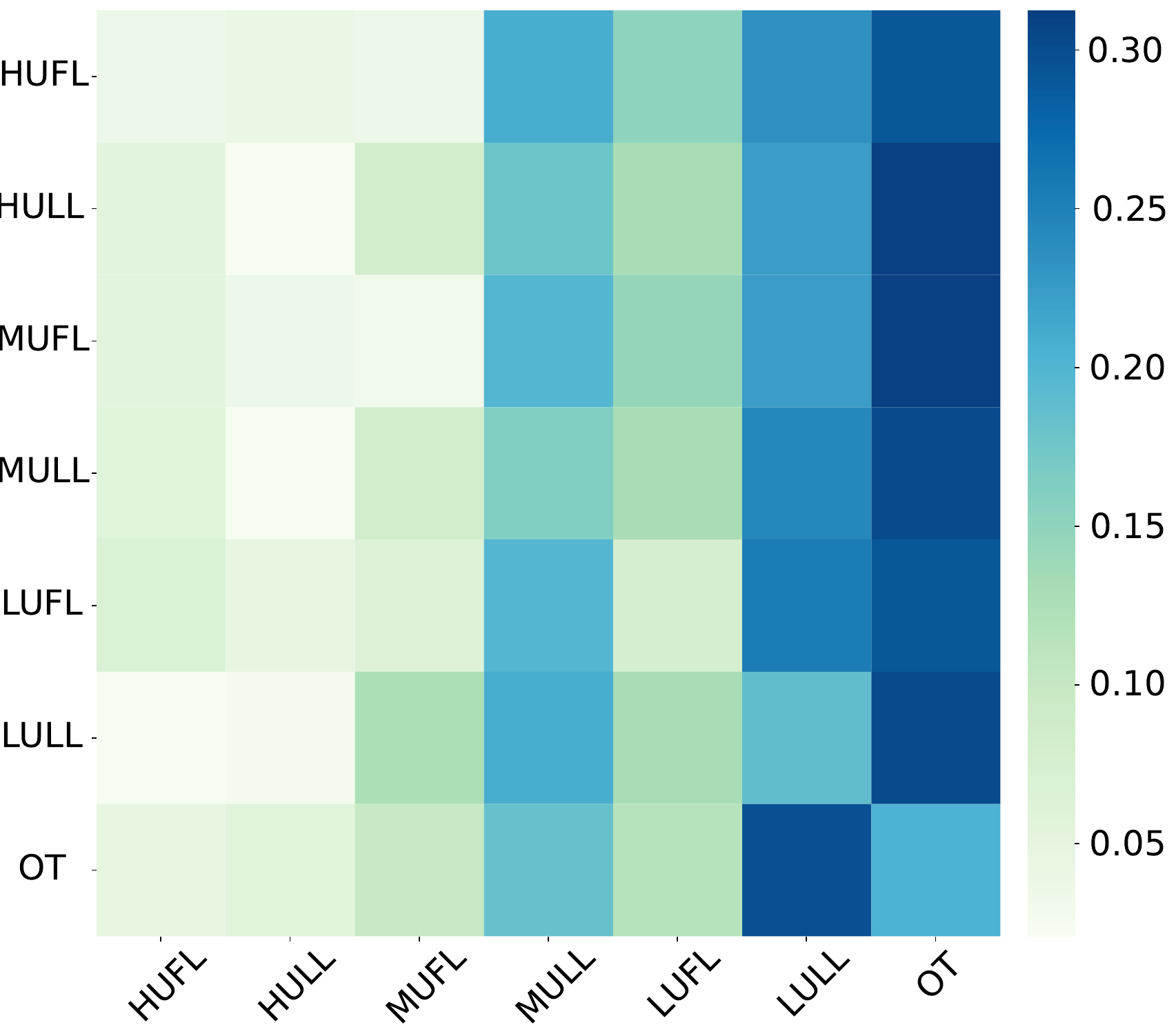}
        \caption*{(b) Prompt encoder.} 
    \end{minipage}
    % \caption{\small Attention maps of Transformer and LLM encoders, \re{capturing complementary variables' pairwise relations}: \re{(a) Transformer captures local and individualized relations for each variable; (b) LLM captures global and shared relations for all multi-variables.}} 
    \caption{\small Attention maps of Transformer and LLM encoders.}
    \label{fig:encoder_att_vis}
\end{figure}
\fi

\subsubsection{Last Token Attention Analysis.}
We visualize the attention of the last token $<\Delta_T>$ from the final layer of GPT-2. First, we segment the words and time series values in the prompt into different segments. Then, we visualize the attention of the last token to the previous segments to verify which part of the last token receives the most attention scores. 
As shown in Fig.~\ref{fig:last_token}: the highest attention from the last token is directed toward the time series value, indicating that \emph{the last token effectively captures the value information of the time series}.
% , which is crucial for efficient forecasting.

\subsubsection{Encoder Attention Analysis.} We visualize the variable attention map from the time series and prompt encoders, respectively, in Fig.~\ref{fig:encoder_att_vis} (a) and (b), \re{each row showing its variable attention to different column variables}. The time series attention is from a Pre-LN Transformer encoder, and the prompt attention is from the LLM. It shows that \emph{\re{Transformer and LLM} capture complementary information of multivariable interrelations: the Transformer time-series attention is local and variable-specific, LLM textual attention is universal and captures global dependencies between variables.} %that can enhance the model's ability to understand data relationships.
In Fig.~\ref{fig:encoder_att_vis} (a), the Transformer attention map %shows how the Transformer encoder distributes its attention across different variables. This 
is local and captures the variable-specific temporal dependencies within the variables.  In Fig.~\ref{fig:encoder_att_vis} (b), the LLM focuses on a broader range of variables, indicating its capability to 
capture global and shared dependencies effectively. 
%Additionally, the LLM attention map helps identify which variables are most important. 
Thus, integrating the LLM with the Transformer enables the TimeCMA to leverage local and global dependencies, enhancing forecasting performance.

% \begin{figure}[t]
% \centering
%     \begin{minipage}[b]{0.495\linewidth}
%         \centering
%         \includegraphics[width=\linewidth]{figures/TS_Embeddings_big.pdf}
%         \caption*{(a) Time series embeddings.}
%     \end{minipage}
%     \hfill
%     \begin{minipage}[b]{0.495\linewidth}
%         \centering
%         \includegraphics[width=\linewidth]{figures/Prompt_Embeddings_big.pdf}
%         \caption*{(b) Prompt embeddings.} 
%     \end{minipage}
%     \hfill
%     \begin{minipage}[b]{0.495\linewidth}
%         \centering
%         \includegraphics[width=\linewidth]{figures/Aligned_Embeddings_big_big.pdf}
%         \caption*{(c) Retrieved embeddings.} 
%     \end{minipage}
%     \hfill
%     \begin{minipage}[b]{0.495\linewidth}
%         \centering
%         \includegraphics[width=\linewidth]{figures/Forecasted_Embeddings_big.pdf}
%         \caption*{(d) Forecasted embeddings.} 
%     \end{minipage}
%     \caption{T-SNE visualization.} 
%     \label{fig:tsne}
% \end{figure}

\begin{figure}[t]
\includegraphics[width=\columnwidth]{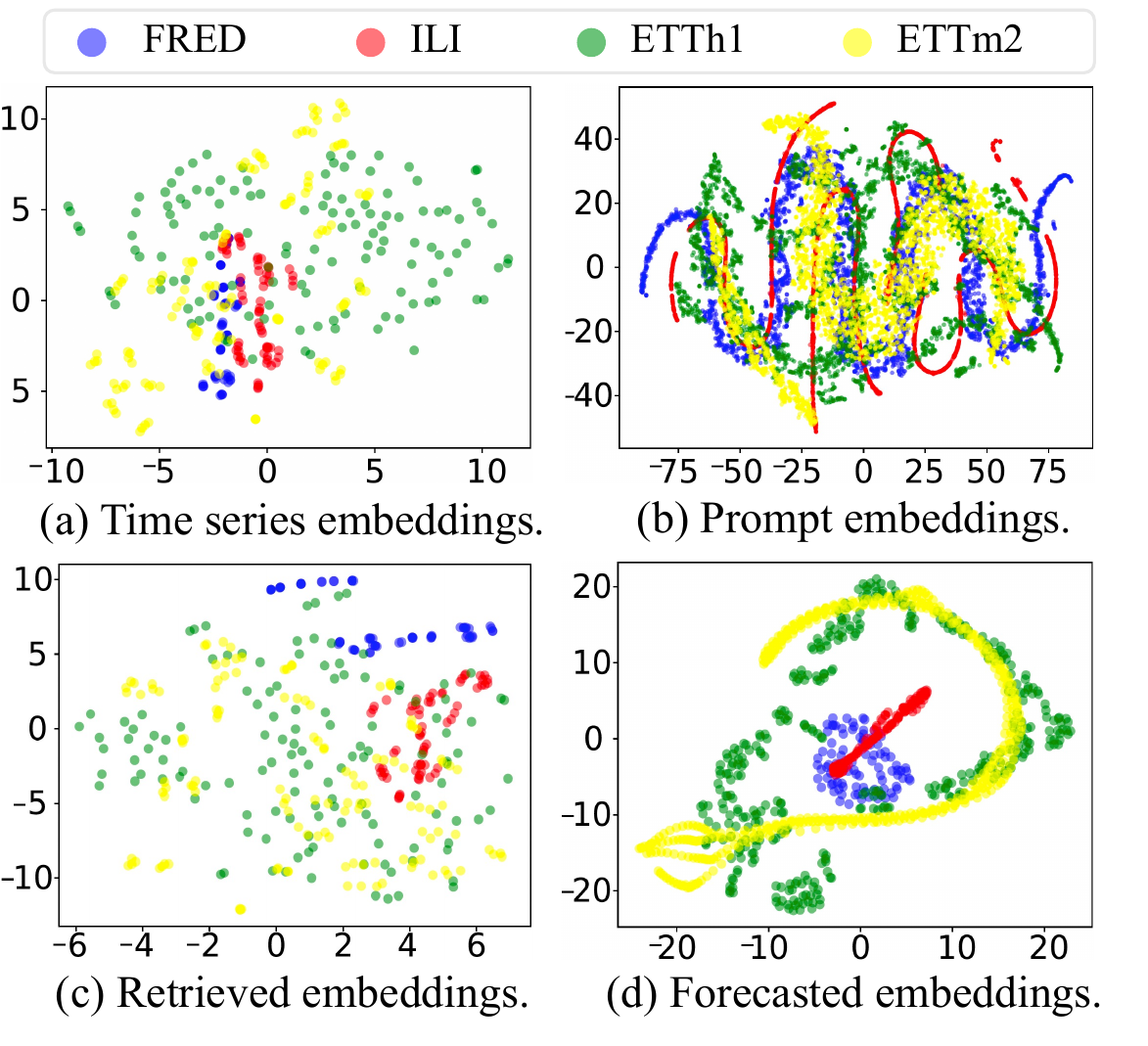} 
    \caption{T-SNE visualization on four datasets.}
    \label{fig:tsne}
\end{figure}

\if 0
\begin{figure}[t]
%\vspace{-0.2cm}
\centering
    \subfigure[TS embeddings.]{
    \includegraphics[width=0.45\linewidth]{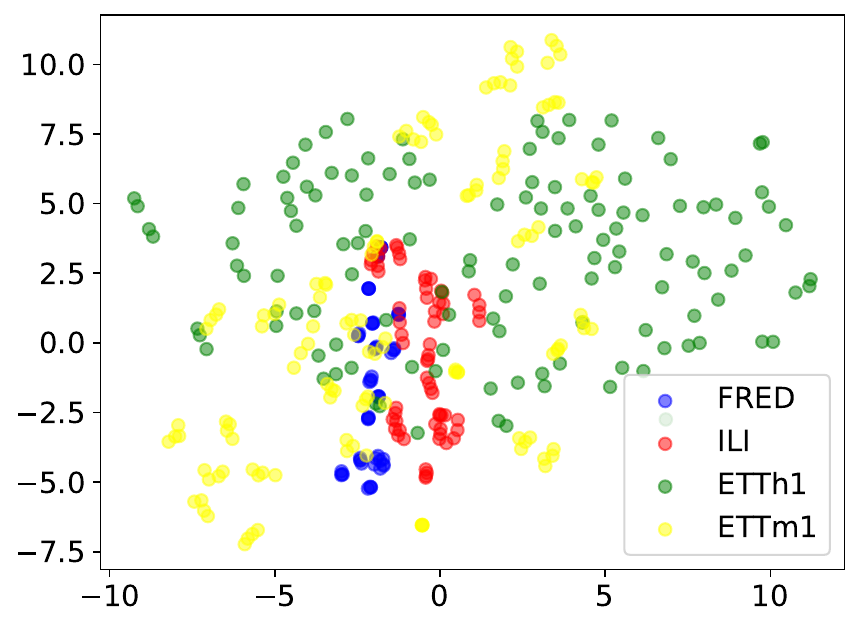}
    }
    \subfigure[Prompt embeddings.]{
    \includegraphics[width=0.45\linewidth]{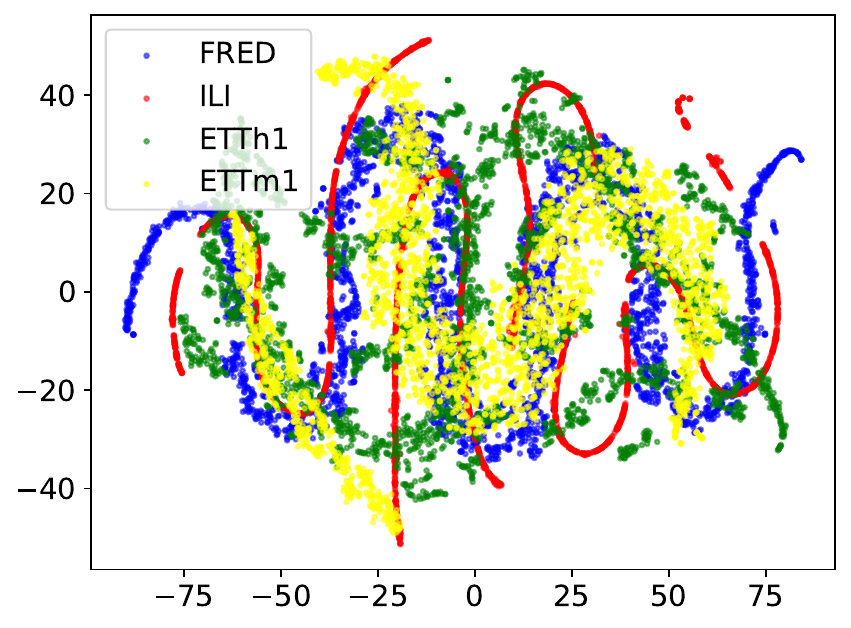}
    }
    \subfigure[Retrieved embeddings.]{
    \includegraphics[width=0.45\linewidth]{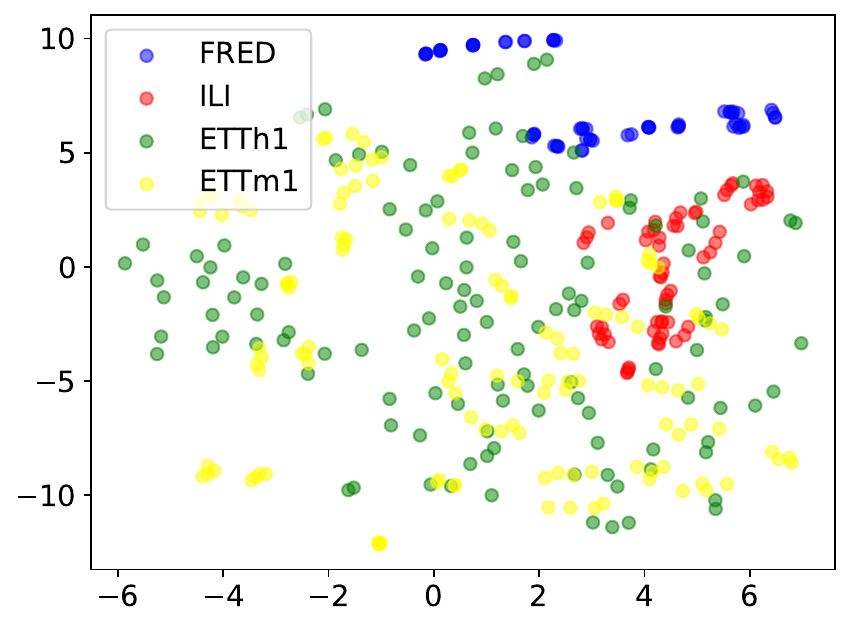}
    }
    \subfigure[Forecasted TS embeddings.]{
    \includegraphics[width=0.45\linewidth]{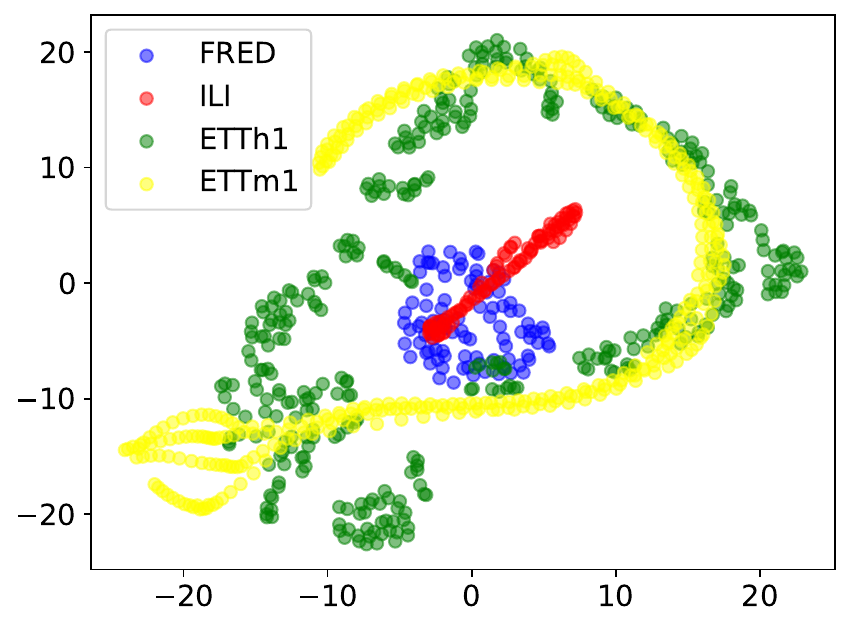}
    }
    % \vspace{-10pt}
    \caption{\small T-SNE visualization.
    %(a) Time series (TS) embeddings from the TS encoder. (b) Prompt embeddings from the prompt encoder. (c) Aligned embeddings from the cross-modality alignment. (d) Forecasted TS embeddings from the projection layer.
    }
    \label{fig:tsne}
    % \vspace{-10pt}
\end{figure}
\fi

\subsubsection{T-SNE Visualization.} 
Fig.~\ref{fig:tsne} presents T-SNE visualization of time series (TS) and prompt embeddings. In Fig.~\ref{fig:tsne} (a), the points are clustered by dataset, indicating that the Transformer captures the specific characteristics of each dataset. Fig.~\ref{fig:tsne} (b) shows that prompt embeddings have more complex inter-relations than TS embeddings. Fig.~\ref{fig:tsne} (c) tightly integrates cross-modality TS embeddings with higher similarity, making the retrieved time series embedding more cohesive. Fig.~\ref{fig:tsne} (d) illustrates that forecasted TS form well-separated clusters for each dataset. This suggests that the projection effectively utilizes the retrieved embeddings to generate accurate forecasts. Overall, the step-by-step refinement shows how the TimeCMA improves data representations.
% \vspace{-0.3cm}
% \vspace{-0.1cm}

\section{Conclusion}
\label{conclusion}
This paper presents TimeCMA, an LLM-empowered framework via cross-modality alignment for multivariate time series forecasting. A cross-modality alignment module is designed to aggregate the time series and LLM branches based on channel-wise similarity retrieval to enhance forecasting. TimeCMA shows promise in using the last token embedding to reduce computational costs and accelerate the inference speed of the LLM-based method. Sufficient experiments offer insights into the efficacy and efficiency of TimeCMA.

\section{Acknowledgments}
This study is supported under the RIE2020 Industry Alignment Fund – Industry Collaboration Projects (IAF-ICP) Funding Initiative, as well as cash and in-kind contributions from the industry partner(s).
\bibliography{aaai25}

\end{document}